\documentclass[acmsmall,screen,nonacm]{acmart}

\usepackage[many]{tcolorbox}    	% for COLORED BOXES (tikz and xcolor included)
\definecolor{main}{HTML}{E3E3E3}    % setting main color to be used

\newtcolorbox{boxM}{
    fontupper = \color{black},
    rounded corners,
    arc = 6pt,
    colback = main!80, 
    colframe = black, 
    boxrule = 1pt, 
    bottomrule = 1pt,
    enhanced,
    width = 0.85\linewidth
}

\usepackage{soul}
\usepackage[normalem]{ulem}
\usepackage{subcaption}

\usepackage{graphicx}
\usepackage{amsmath}
\usepackage{booktabs}
\usepackage{algorithmic}
\usepackage[ruled,vlined,linesnumbered]{algorithm2e}
\usepackage{caption}

\AtBeginDocument{%
  \providecommand\BibTeX{{%
    \normalfont B\kern-0.5em{\scshape i\kern-0.25em b}\kern-0.8em\TeX}}}

\begin{document}
% \setcopyright{none}
% \settopmatter{printacmref=false}

\title{Out-of-Distribution Data: An Acquaintance of Adversarial Examples - A Survey}

\author{Naveen Karunanayake}
\email{naveen.karunanayake@sydney.edu.au}
% \orcid{0002-7081-2958}
\affiliation{%
  \institution{The University of Sydney}
  \country{Australia}
}

\author{Ravin Gunawardena}
\affiliation{%
  \institution{The University of New South Wales}
  \country{Australia}}
\email{r.gunawardena@unsw.edu.au}

\author{Suranga Seneviratne}
\affiliation{%
  \institution{The University of Sydney}
  \country{Australia}}
  \email{suranga.seneviratne@sydney.edu.au}

\author{Sanjay Chawla}
\affiliation{%
  \institution{Qatar Computing Research Institute, HBKU}
  \country{Qatar}}
\email{schawla@hbku.edu.qa}

\begin{abstract}

Deep neural networks (DNNs) deployed in real-world applications can encounter out-of-distribution (OOD) data and adversarial examples. These represent distinct forms of distributional shifts that can significantly impact DNNs' reliability and robustness. Traditionally, research has addressed OOD detection and adversarial robustness as separate challenges. This survey focuses on the intersection of these two areas, examining how the research community has investigated them together. Consequently, we identify two key research directions: robust OOD detection and unified robustness. Robust OOD detection aims to differentiate between in-distribution (ID) data and OOD data, even when they are adversarially manipulated to deceive the OOD detector. Unified robustness seeks a single approach to make DNNs robust against both adversarial attacks and OOD inputs. Accordingly, first, we establish a taxonomy based on the concept of distributional shifts. This framework clarifies how robust OOD detection and unified robustness relate to other research areas addressing distributional shifts, such as OOD detection, open set recognition, and anomaly detection. Subsequently, we review existing work on robust OOD detection and unified robustness. Finally, we highlight the limitations of the existing work and propose promising research directions that explore adversarial and OOD inputs within a unified framework.

\end{abstract}




\maketitle

\section{Introduction}
\label{sec:introduction}

Early stages of deep learning were heavily dependent on a closed-world assumption~\cite{closed_world_1,closed_world_2}. That is, a model operates under the premise that it would only encounter inputs conforming to the same distribution as its training data during inference. However, in real-world deployments, the model is certain to encounter inputs from a wide range of unseen instances that exhibit various distributional shifts~\cite{distributional_shifts}. 
In such situations, deep learning models tend to produce overconfident, yet incorrect predictions  
regardless of their unfamiliarity, making them unsuitable for many applications~\cite{nguyen2015deep}. For instance, in safety-critical scenarios such as autonomous driving, the driving system must promptly alert the drivers and transfer control when it detects unfamiliar scenes or objects not encountered during its training phase~\cite{nitsch2021out}. Therefore, it is essential to build robust models that can detect out-of-distribution (OOD) samples and make correct predictions for in-distribution (ID) samples.

A distribution shift can occur in two main types. One type arises from changes in the inherent meaning of the data. Specifically, within the context of deep learning classifiers, a shift in meaning would indicate the presence of data associated with novel, previously unseen classes. This shift is often referred to as a \textit{semantic shift}~\cite{hendrycks2016baseline}. The vast majority of work in out-of-distribution detection~\cite{hendrycks2016baseline,liang2017enhancing,react,dice} focuses on identifying semantic shifts. The second form of distributional shifts stems from subtle modifications applied to the data, such as adversarial perturbations~\cite{fgsm}, style changes~\cite{style_changes}, and domain shifts~\cite{quinonero2008dataset}. These modifications, which are collectively referred to as \textit{covariate shifts}~\cite{covariate_1,covariate_2}, induce variations in the variance of the data distribution. 
Adversarial perturbations, while falling under the category of covariate shifts, stand apart from others due to their malicious intent in crafting imperceptible alterations. 

\begin{figure}[t]
\centering
\includegraphics[width=0.55\textwidth]{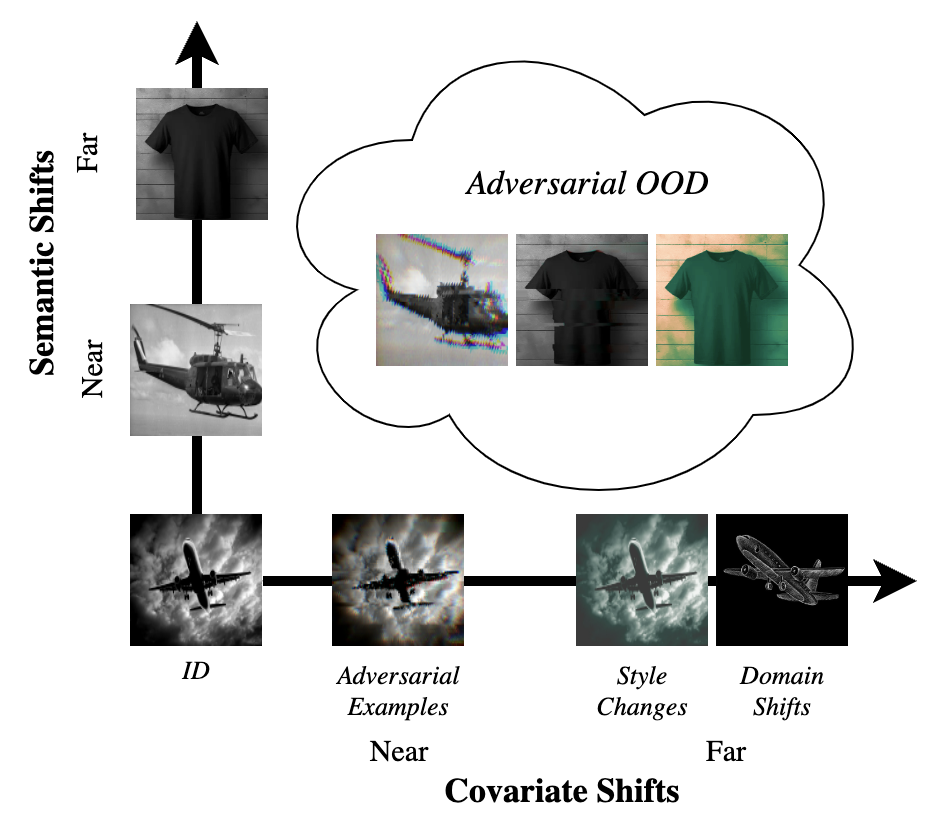} 
\caption{\centering Types of distributional shifts. \break The x and y axes represent the different forms of covariate and semantic shifts, respectively. }
\label{Fig:OOD spectrum}
\end{figure}

We illustrate these different forms of distributional shifts in Figure~\ref{Fig:OOD spectrum}, highlighting the extent of deviation from the training data. The x and y axes represent the different types of covariate shifts and semantic shifts, respectively. On the x-axis, adversarial examples remain closest to ID data, as they involve introducing imperceptible perturbations to clean ID data. Moving along the x-axis, we encounter style changes and domain shifts, which can arise from noticeable changes in the appearance of ID data. These covariate shifts demonstrate the potential variations within an ID class. Along the y-axis, we find semantic shifts, which stem from previously unseen classes during the training phase. Semantic shifts can be classified into two distinct types: near OOD and far OOD, depending on the proximity of an unseen class to a known ID class~\cite{yang2022openood}. At the intersection, there lies adversarial OOD, where OOD inputs undergo adversarial perturbations. In this survey, we focus on the inter-relatedness of adversarial examples and OOD data. Even though the vast majority of the existing work operates in silos, focusing either on OOD detection or adversarial robustness, some of the emerging work has investigated OOD data and adversarial inputs in a unified framework and proposed methods even to detect them together~\cite{lee2022gradient,lee2018simple}. Since real-world systems can receive both adversarial and OOD data as inputs, whether occurring naturally or with adversarial intent, it is important to analyse how these inputs collectively pose threats to machine learning systems and explore strategies to mitigate them effectively. To elaborate further, we next describe some applications where adversarial and OOD inputs co-occur in the real world.

\noindent{\textbf{Financial fraud}. 
Machine learning models are used to detect fraudulent transactions. Adversarial examples could be used to manipulate transactions in a way that evades fraud detection systems~\cite{fursov2021adversarial}. Additionally, OOD data is relevant to detect new and emerging forms of financial fraud that may not have been encountered previously~\cite{hilal2022financial}}. \\ \vspace{-2mm}

\noindent{\textbf{Cybersecurity}. Adversarial attacks can bypass traditional security systems, such as spam filters or malware detectors. Attackers may craft emails or malicious files that evade detection by adding subtle perturbations, or encoding techniques that make the content appear benign to the classifier~\cite{aryal2111survey}. OOD data is also relevant in cybersecurity, where identifying novel malware or cyber threats such as zero-day attacks~\cite{zero_day_attacks} that have not been seen before is crucial to staying ahead of rapidly evolving cyberattacks.} \\ \vspace{-2mm} 

\noindent{\textbf{Autonomous driving.} 
Deep Neural Networks (DNNs) are extensively utilised for image recognition tasks to identify objects on the road. These models are vulnerable to adversarial inputs that can deceive the classifier, posing a significant safety risk~\cite{kong2020physgan}. Moreover, the presence of OOD data, such as rare and novel objects not encountered during training, can cause the DNN to make erroneous predictions with high confidence, potentially leading to dangerous driving decisions~\cite{nitsch2021out}. For instance, in May 2016, a Tesla in autopilot mode caused the first fatality associated with autonomous driving. Subsequent analysis found that the computer vision-assisted navigation system misclassified the white side of a trailer as the bright sky, resulting in a collision near Williston, Florida, USA. Tragically, the collision led to the death of the car's driver due to sustained injuries, as detailed in the report by the National Transportation Safety Board~\cite{NTSB2016TeslaCrash}. Here, the white side of the trailer, appearing as the bright sky, is an example of a natural adversarial example occurring in the physical world~\cite{kurakin2018adversarial}.} \\ \vspace{-3mm}

Accordingly, adopting a single technique to strengthen a model's robustness against both adversarial and OOD inputs, termed henceforth as \textit{unified robustness}, is important as it yields several benefits in real-world systems. Primarily, such an approach can streamline the implementation process and reduce the computational overhead. Therefore, instead of incorporating two separate defence mechanisms, a unified approach reduces the system complexity and resource consumption, making it more practical for real-world deployment. This also conforms to the design principle ''Economy of Mechanism''~\cite{design_principles}, which emphasises that the simplest and smallest solutions are the easiest to secure.
Moreover, from a practical standpoint, a unified technique enables easier maintenance and updates. With a single defence strategy, developers can focus on refining and enhancing a unified model, leading to more efficient updates and continuous improvement over time. This significantly reduces the maintenance burden compared to managing and updating separate defence mechanisms for adversarial and OOD detection. 
In addition, with the recent introduction of AI regulations~\cite{whitehouseBlueprintBill,artificialintelligenceactHome}, the emphasis on ensuring model \textit{robustness} and \textit{explainability} has increased. This implies that models should be able to provide a reason for their predictions. A unified approach provides more convenience in explaining the outcome of a model than two distinct methods. Furthermore, unifying the two defences allows for complementary benefits, as adversarial examples and OOD data may share similarities in their analogous patterns and distributions. By jointly considering these characteristics, a unified technique can better leverage shared knowledge and can have mutual reinforcement effects to improve the overall detection accuracy for both types of threats.

\begin{figure}[t]
\centering
\includegraphics[width=0.55\textwidth]{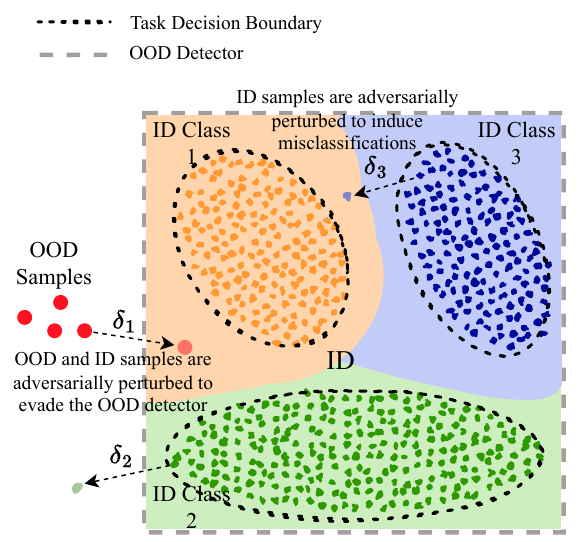} 
\caption{\centering Adversarial examples and OOD detection. \break The coloured regions represent the regions corresponding to different ID classes, and the margin represents the model's decision boundary.} 
\label{Fig:IntroFigure}
\end{figure}

Another line of work that intersects OOD and adversarial inputs is \textit{robust OOD detection} which refers to handling of adversarial inputs attempting to evade the OOD detector. As shown in Figure~\ref{Fig:IntroFigure} adversarial examples can either target the main classification problem (i.e., $\delta_3$) or the OOD detector. Robust OOD detection differs from the conventional problem of adversarial robustness, as it focuses on handling adversarial inputs trying to evade the OOD detector, which can lead to misclassification of OOD inputs as ID (i.e., $\delta_1$), or vice versa (i.e., $\delta_2$). Moreover, Robust OOD detection also differs from conventional OOD detection, as it performs standard OOD detection even in the presence of adversarial manipulations. In this work, we survey how OOD and adversarial inputs contribute to strengthening the overall robustness of DNNs through the means of Robust OOD detection and Unified robustness.

Several surveys have been conducted over the past years in the areas of adversarial example detection and defences~\cite{adversarial_survey,liang2022adversarial}, OOD detection and generalisation~\cite{detection_survey,salehi2021unified,cui2022out,ghassemi2022comprehensive,generalization_survey}. These surveys have contributed significantly to understanding the techniques and challenges associated with each domain. To our knowledge, our survey is the first to analyse the intersection between adversarial examples and OOD detection, filling a literature gap by reviewing state-of-the-art techniques for robust OOD detection and unified robustness. More specifically, we make the following contributions in this paper.

\begin{itemize}
    \item We investigate adversarial examples and OOD data in a unified framework and systematically review two topics on their intersection.

    \item We analyse state-of-the-art techniques for unified robustness and robust OOD detection, identifying similarities and differences and providing insights into their strengths and weaknesses.
    
    \item We bring attention to the limitations of existing robust OOD detection and unified robustness techniques and conclude the survey by discussing open challenges and future research directions in these areas.
\end{itemize}

The rest of the paper is organised as follows. In Section~\ref{sec:background}, we provide the background information, while Section~\ref{sec:taxonomy} introduces the novel taxonomy of existing work at the intersection of adversarial examples and OOD detection. Section~\ref{sec:experimental_details} discusses the experimental details commonly encountered across the literature on robust OOD detection and unified robustness. We provide a detailed analysis of the existing work on robust OOD detection and unified robustness in Sections~\ref{sec:robust_ood_detection} and~\ref{sec:unified_robustness}, respectively, and discuss the limitations of these methods as well as potential future research directions in Section~\ref{sec:limitations_future}. Finally, Section~\ref{conclusion} concludes the paper.

\section{Background} \label{sec:background}

In this section, we introduce the terminology and the formal definitions, along with other preliminaries related to robust OOD detection and unified robustness in DNNs. 

\subsection{Deep neural networks}

Deep neural networks (DNNs) can be represented mathematically as a composition of multiple layers of functions, where each layer consists of neurons or nodes trained for a particular task, such as classification and regression. 
For an input $x \in \mathcal{D}_{\text{train}}$, where $\mathcal{D}_{\text{train}}$ is the training dataset, a DNN can be mathematically represented as,
\begin{equation}
   f(x;\theta) = F_n(F_{n-1}(...(F_2(F_1(x))))).
\end{equation}
Here, $\theta$ denotes the parameters of the network, and $F_i$ is a task-specific differentiable transformation layer. The DNN is trained to minimise the error between the prediction $f(x;\theta)$ and the ground truth $y$ over the training set by minimising a differentiable loss function $\mathcal{L}(f(x;\theta),y)$. 
\begin{equation}
    \label{eq:lossoptimization}
    \theta^* = \underset{\theta}{\arg\min} \sum_{i \in \mathcal{D}_{\text{train}} } \mathcal{L}(f(x_i;\theta), y_i)
\end{equation}

\subsection{Out-of-distribution detection}

Suppose the learning task is defined on a domain $\mathcal{X}$, and all data instances of $\mathcal{D}_{\text{train}}$ are drawn from a specific data distribution, referred to as $\mathcal{D}_{\text{in}} \subset \mathcal{X}$. Nevertheless, during testing, DNNs encounter samples from distributions beyond $\mathcal{D}_{\text{in}}$. Therefore, the objective of an OOD detector $g(x)$ can be formally defined as,
\begin{equation}
\label{eq:ood_detection}
 g(x)= 
    \begin{cases}
        1 & \text{if } x \in \mathcal{D}_{\text{in}}, \\
        0 & \text{else} .
    \end{cases}
\end{equation}
Maximum softmax probability (MSP)\cite{hendrycks2016baseline}, Mahalanobis distance score (MDS)\cite{lee2018simple}, and Energy~\cite{energy} are among the simpler OOD detectors, while DICE~\cite{dice}, ReAct~\cite{react}, and LogitNorm~\cite{logitnormalization} represent more complex approaches to OOD detection. The OpenOOD leaderboard,\footnote{https://zjysteven.github.io/OpenOOD/} developed by Zhang et al.~\cite{zhang2023openood}, 
provides a comprehensive benchmark of OOD detection methods over the years, showcasing the current state-of-the-art in this field.

\begin{figure}[t]
\centering
\includegraphics[width=0.48\textwidth]{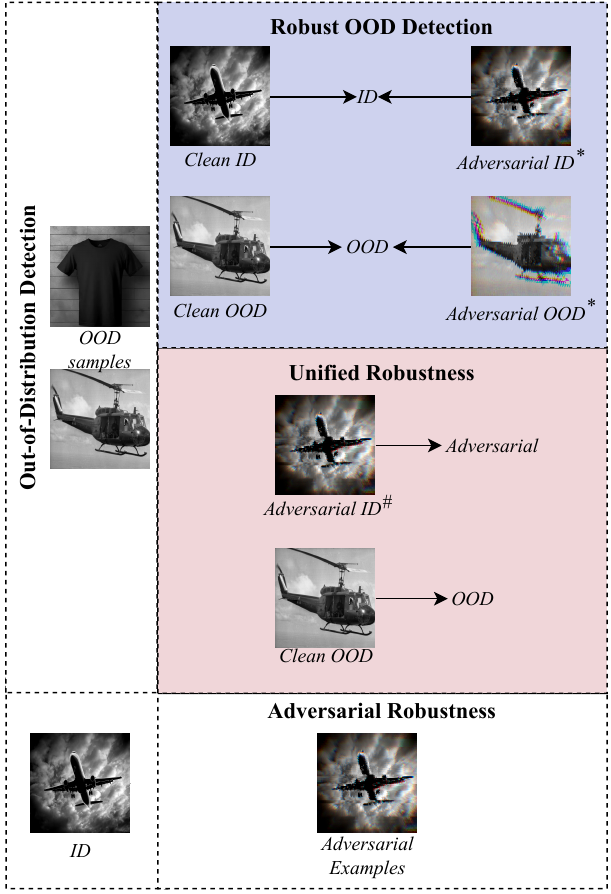} 
\caption{\centering Visualisation of robust OOD detection and unified robustness in relation to standard OOD detection and adversarial robustness. \break ({\bf Note} - * indicates adversarial examples attempting to evade the OOD detector, and \# indicates adversarial examples aimed at deceiving the primary classifier.) }
\label{Fig:task_connection}
\end{figure} 

\subsection{Adversarial robustness}
\label{subsec:adv_ex_detection}
Adversarial examples~\cite{szegedy2013intriguing} are maliciously crafted inputs to deceive DNNs by adding imperceptible perturbations to original data. An adversarial example $x'$ for a given input $x$ is formally defined as,
\begin{align}
\label{eq:adv_example}
\begin{aligned}
    &x' = x + \delta ; \  ||\delta||_p \leq \epsilon, \\
    &\text{s. t.} \  f(x') \neq f(x) .
    \end{aligned}
\end{align}

Here, $\delta$ represents a carefully crafted perturbation, constrained by a maximum $p$-norm of $\epsilon$ where $p \in \{1,2,\infty\}$. Various techniques for generating adversarial inputs propose distinct approaches to solve Equation~\ref{eq:adv_example}~\cite{fgsm,madry2017towards,carlini2017towards,kurakin2018adversarial,dong2018boosting}. On the other hand, adversarial robustness entails either the capability to maintain accurate predictions in the presence of adversarial perturbations or the ability to detect and mitigate the impact of such malicious inputs that could otherwise degrade the performance of DNNs. A model is said to be adversarially robust if,
\begin{equation}
\label{eq:adversarial_robustness}
    \forall ||\delta||_p \leq \epsilon,   \  f(x+\delta) = f(x).
\end{equation}
Adversarial training~\cite{fgsm,madry2017towards}, as shown in Equation~\ref{eq:pgd_adv_training}, is widely recognised as one of the most commonly employed empirical defences against adversarial examples.
\begin{equation}
\label{eq:pgd_adv_training}
\min_{\theta} \mathbb{E}_{(x, y) \sim \mathcal{D_{\text{train}}}} \left[ \max_{\| \delta \|_p \leq \epsilon} \mathcal{L}(f(x+\delta;\theta), y) \right]
\end{equation}
The progress over the past decade and the current state-of-the-art in adversarial robustness have been reported in the RobustBench library.\footnote{https://robustbench.github.io/}

\begin{figure*}[t]
\centering
\includegraphics[width=0.95\textwidth]{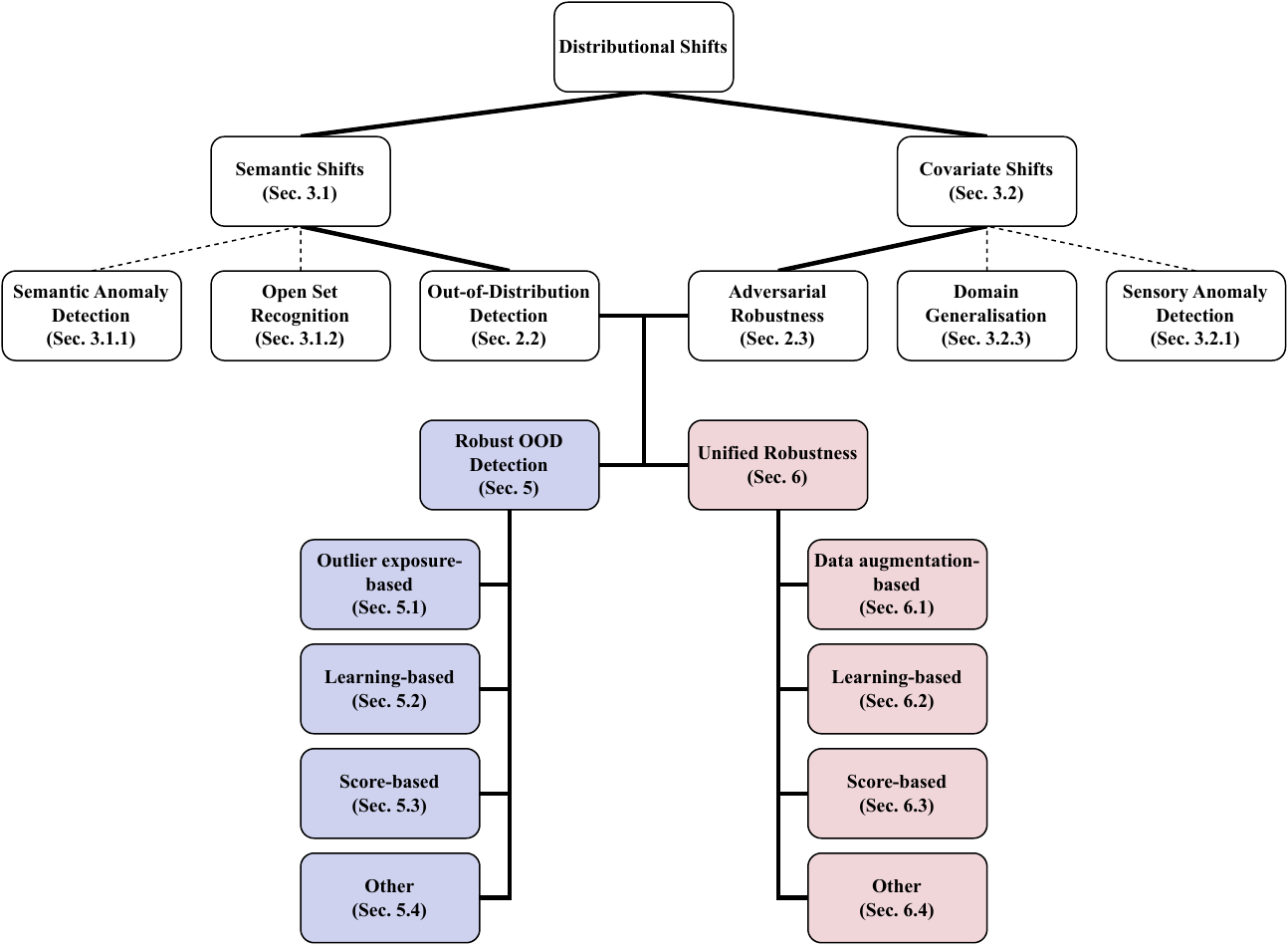} 
\caption{\centering The proposed taxonomy based on different forms of distributional shifts. \break Primarily, distributional shifts are categorised into semantic and covariate shifts. Under semantic shifts, we encounter OOD inputs, while adversarial examples fall under covariate shifts. Our focus lies at the intersection of these two categories, specifically on robust OOD detection and unified robustness.} 
\label{fig:taxonomy}
\end{figure*}

\subsection{Robust OOD detection}
\label{subsec:robustood}
In the context of robust OOD detection, the adversarial objective is to bypass the OOD detector $g(x)$ instead of the primary classifier $f(x)$ in Equation~\ref{eq:adv_example}~\cite{azizmalayeri2022your,chen2020robust,fort2022adversarial}. Therefore, a robust OOD detector should be able to capture whether an input is OOD or not, even if it is adversarially perturbed, as defined in Equation~\ref{eq:robust_ood_detection}.  
\begin{equation}
\label{eq:robust_ood_detection}
\forall ||\delta||_p \leq \epsilon , \ g(x+\delta)= 
    \begin{cases}
        1 & \text{if } x \in \mathcal{D}_{\text{in}}, \\
        0 & \text{else} .
    \end{cases}
\end{equation}
Similar techniques, as listed in Section~\ref{subsec:adv_ex_detection}, can be employed for generating adversarial inputs against the OOD detector. In this case, the optimization problem is related to the OOD detector function instead of the primary classification loss function. For instance, gradient-based adversarial example generation techniques are inclined to use \textbf{gradients of the OOD score} with respect to the inputs rather than \textit{gradients of the loss} with respect to the inputs~\cite{fort2022adversarial}.

\subsection{Unified robustness}
\label{subsec:unifiedrobust}
We define the term \textit{unified robustness} to encompass methods that either make the model robust against both OOD and adversarial inputs simultaneously or detect both OOD and adversarial inputs following the same approach. Here, it is important to note that adversarial examples target the main classification problem but not the OOD detector, as in robust OOD detection. In most cases, unified robust methods achieve adversarial robustness as in Equation~\ref{eq:adversarial_robustness}, while increasing the separability in the embedding space between ID and OOD data. This functionality of a unified robust method ($\mathcal{UR}$) can be mathematically represented as,
\begin{equation}
\label{eq:unifiedrobustness}
\mathcal{UR} : \begin{cases}
    \forall ||\delta||_p \leq \epsilon,   \  f(x+\delta) = f(x) \\

   g(x) =  \begin{cases}
        1 & \text{if } x \in \mathcal{D}_{\text{in}}, \\
        0 & \text{else} .
    \end{cases}
    \end{cases}
\end{equation}

Different from the above definition, previous work has also shown the capability of \textit{detecting} both OOD and adversarial inputs using the same method but with different sets of parameters (e.g., weights, biases, and thresholds)~\cite{lee2022gradient,lee2018simple}.

Figure~\ref{Fig:task_connection} visualises the tasks of robust OOD detection and unified robustness in relation to standard OOD detection and adversarial robustness. Specifically, robust OOD detection aims to accurately identify both clean ID and OOD inputs, as well as adversarial ID and OOD inputs crafted to evade the OOD detector. In contrast, unified robustness focuses on identifying adversarial ID inputs designed to deceive the primary classifier and clean OOD inputs that should not be considered for inference. In the following section, we introduce our taxonomy, placing robust OOD detection and unified robustness at the intersection of OOD detection and adversarial robustness.

\section{Taxonomy}
\label{sec:taxonomy}

We systematically categorise the existing work on robust OOD detection and unified robustness, providing a clear context on how they are positioned relative to other related ideas. Figure~\ref{fig:taxonomy} visualises our proposed taxonomy, which is structured around the fundamental concept of distributional shifts. \textit{Distributional shifts} occur when the underlying data distribution deviates from the distribution encountered during model training. Based on the nature of these shifts, the taxonomy branches out into two categories as semantic and covariate shifts. 

\subsection{Semantic shifts}
As discussed in Section~\ref{sec:introduction}, semantic shifts refer to situations where new, previously unseen classes emerge in the data. We classify semantic shifts into three main categories; i) semantic anomaly/outlier detection, ii) open set recognition, and iii) out-of-distribution detection, based on their respective objectives. 

\subsubsection{\textbf{Semantic anomaly/outlier detection}}
Semantic Anomaly/Outlier Detection~\cite{detection_survey,chandola2009anomaly} aims to identify abnormal or anomalous data points that deviate significantly from the ID classes. It focuses on detecting rare instances that exhibit distinct characteristics that differ from the established properties of the known ID classes. \\ \vspace{-2mm}

\noindent\textbf{Example}: Anomalies in a computer network's traffic pattern might suggest that a compromised computer is transmitting sensitive data to an unauthorized destination. Similarly, anomalous MRI images could signal the presence of malignant tumours. In credit card transaction data, anomalies may indicate instances of credit card or identity theft~\cite{chandola2009anomaly}.

\subsubsection{\textbf{Open set recognition}}
Open Set Recognition~\cite{open_set_survey,wang2020high} approaches focus on recognising novel classes that were not present during the model's training. Here, the objective is two-fold: to ensure that the model accurately classifies instances from known classes while also correctly identifying instances from previously unseen classes by flagging them as unknown. Open set approaches leverage a single classifier to achieve this dual objective. It is important to note that these methods implicitly assume that the distribution shift between known and unknown classes is relatively smaller, not at the anomaly level. \\ \vspace{-2mm}

\noindent\textbf{Example}: In website fingerprinting~\cite{wang2020high,dahanayaka2023robust}, Open Set Recognition involve training a classifier to recognise known website traffic patterns (e.g., common websites like Google, Facebook, etc.). During training, the classifier learns to identify features unique to these known websites. Then, during testing, when encountering traffic from previously unseen websites (unknown classes), the classifier should ideally recognise them as unknown rather than misclassifying them as one of the known websites.

\subsubsection{\textbf{Out-of-distribution detection}}
Out-of-Distribution Detection~\cite{detection_survey} methods aim to identify samples lying outside the known distribution of the training data. The primary objective is to distinguish between in-distribution data (i.e., data from the training distribution) and out-of-distribution data, which might originate from a different source or context. Accordingly, OOD detection encompasses a broad spectrum of unseen classes, ranging from those that are semantically similar to highly different anomalies. Therefore, it can be viewed as addressing the generalised semantic shift detection problem, which includes elements of both semantic anomaly detection and open set recognition.

\noindent\textbf{Example}: If an autonomous vehicle is trained on data from urban environments and suddenly encounters a challenging off-road terrain, out-of-distribution detection would help the system recognise that the current scenario is different from its training conditions, prompting it to adapt its decision-making process.

\subsection{Covariate shifts}
Covariate shifts result due to alterations in the variance of the data distribution, reflecting changes in the statistical properties of input features. These changes may include shifts in the scale, style, or patterns of the data. We further divide this branch into three subcategories: i) sensory anomaly/outlier detection, ii) adversarial robustness, and iii) domain generalisation. Although the latter two techniques share the goal of enhancing model generalisability, they vary in their response to the \textbf{malicious intent} behind the covariate shifts they address.

\subsubsection{\textbf{Sensory anomaly/outlier detection}}
\label{subsubsec:sensory_anomaly_detection}
Sensory Anomaly/Outlier Detection~\cite{detection_survey} focuses on identifying data points \textbf{within a known ID class} that appear unusual due to significant differences in their sensory attributes, such as appearance or quality. \\ \vspace{-2mm} 

\noindent\textbf{Example}: In the context of sensors on an autonomous vehicle, sensory anomaly detection could come into play if one of the LiDAR sensors starts providing irregular data, signalling a potential malfunction or damage to that specific sensor.

\subsubsection{\textbf{Adversarial robustness}}
Adversarial Robustness~\cite{silva2020opportunities} focuses on enhancing the model's resilience against adversarial perturbations. These perturbations are meticulously crafted, often based on the model's information, to mislead the model and cause it to make incorrect predictions. Adversarial robustness techniques aim to make the model more resistant to such deceptive inputs, ensuring robustness even when facing more sophisticated attacks. \\ \vspace{-2mm} 

\noindent\textbf{Example}: In the presence of adversarial conditions, such as someone strategically placing confusing road signs or markings, adversarial robustness ensures that the self-driving car maintains accurate navigation decisions despite attempts to mislead the system.

\subsubsection{\textbf{Domain generalisation}}
Domain Generalisation~\cite{domain_generalisation_survey} approaches aim to develop models that can generalise their predictions across different domains that are non-adversarial in nature. These domains could include variations like style changes (i.e., colour, font, background), scene/viewpoint changes, or sketches/cartoons, which are commonly encountered in real-world data. The goal is to create models that can perform well even on data from previously unseen domains, making them more adaptable and versatile in handling diverse data variations. It is worth noting that this differs from sensory anomaly detection, which aims to identify data points deviating significantly from expected patterns within a single domain ({\bf cf.} Section~\ref{subsubsec:sensory_anomaly_detection}).\\ \vspace{-2mm}

\noindent\textbf{Example}: If a self-driving car is trained in one city but adapts to perform well when deployed in a different city with slightly different traffic patterns and road conditions, it demonstrates domain generalisation by effectively applying its learned driving behaviours to a new yet related domain.

\subsection{Intersection of out-of-distribution detection and adversarial robustness}
As we focus on the intersection of OOD detection and adversarial robustness, we survey the works that consider both OOD and adversarial inputs in a unified framework. As can be seen in Figure~\ref{fig:taxonomy}, we identify two lines of work that are motivated by both out-of-distribution detection and adversarial robustness, as robust OOD detection and unified robustness. 

\subsubsection{\textbf{Robust OOD detection}}
\label{subsubsec:robustooddetection}
Robust OOD Detection involves developing techniques capable of identifying and handling OOD inputs, even when they are subjected to adversarial modifications or perturbations aimed at evading the detector ({\bf cf.} Section~\ref{subsec:robustood}). Based on their underlying methodology, we classify the previous work on robust OOD detection into four main branches.  \\ \vspace{-2mm}

\noindent\textbf{i. Outlier exposure-based} methods~\cite{chen2020robust,chen2021atom,hein2019relu} integrate outliers into the training objective of the primary classifier to enhance its resilience against OOD inputs encountered during testing. The core idea is to train the model to produce highly confident predictions for regular training data while yielding uniform predictions (i.e., across classes) for outlier inputs, even in the presence of adversarial perturbations. These techniques employ either an auxiliary dataset or strategically crafted data samples through generators as outliers during model training. \\ \vspace{-2mm} 

\noindent\textbf{ii. Learning-based} methods~\cite{khalid2022rodd,berrada2021make} adopt a specific training approach to enhance the effective separation of ID and OOD data. Unlike outlier exposure-based methods, this line of work does not utilise outlier data during training. It encompasses techniques such as self-supervised learning and pre-training approaches, all aimed at reinforcing the robustness of OOD detection. \\ \vspace{-2mm} 

\noindent\textbf{iii. Score-based} methods~\cite{fort2022adversarial,azizmalayeri2022your} operate by assigning a score that measures how much an input deviates from the training distribution. This score can manifest in various forms, such as distance, probability, or error. Subsequently, the score is compared to a predefined threshold to determine whether the input is OOD. While these methods primarily serve as post-hoc detectors, they are also commonly employed as post-processors in conjunction with training-based methods. \\ \vspace{-2mm}  

\noindent\textbf{iv. Other} category represents a diverse set of approaches that are closely linked to robust OOD detection yet do not directly fit into any of the above categories~\cite{yoon2022evaluating,kaya2022generating,ojaswee2023benchmarking}. For instance, Kaya et al.~\cite{kaya2022generating} proposed an adaptive attack to evade statistical detectors, highlighting the requirement of rigorous evaluation strategies to ensure the adversarial robustness of OOD detectors.

\subsubsection{\textbf{Unified robustness}}
\label{subsubsec:unifiedrobustness}
Unified Robustness methods handle both adversarial and OOD inputs through a single approach. These methods either detect such distributional shifts using post-training mechanisms or enhance the model's robustness against these shifts during training (cf. Section~\ref{subsec:unifiedrobust}). Similar to Section~\ref{subsubsec:robustooddetection}, we classify the work on unified robustness into four categories as follows.  \\ \vspace{-2mm} 

\noindent\textbf{i. Data augmentation-based} methods~\cite{hendrycks2022pixmix,azizmalayeri2023data,pinto2022using} improve the model's robustness against adversarial and OOD inputs by increasing the diversity of the training dataset by creating new data points from existing ones. For instance, interpolated samples are generated using a pair of inputs in the training set and used to regulate the model training process~\cite{pinto2022using}.  \\ \vspace{-2mm}  

\noindent{\textbf{ii. Learning-based}}~\cite{hendrycks2019using_self_supervised,hendrycks2019using_pre-trained,malinin2019reverse} methods come into play in the model training phase. They either adopt more complex learning approaches such as self-supervised learning~\cite{hendrycks2019using_self_supervised} or adopt sophisticated objective functions~\cite{malinin2019reverse} to improve performance on uncertain inputs such as adversarial or OOD inputs. These approaches also closely align with the existing work discussed in the corresponding section under robust OOD detection (cf. Section~\ref{subsubsec:robustooddetection} \textbf{ii}). \\ \vspace{-2mm} 

\noindent\textbf{iii. Score-based} methods~\cite{lee2018simple,kaur2022idecode,osada2023out,qiu2022detecting} employs a score that differentiates clean ID inputs from adversarial/OOD inputs. This category consists of similar approaches to score-based methods discussed under robust OOD detection (cf. Section~\ref{subsubsec:robustooddetection}-\textbf{iii}). For instance, Kaur et al.~\cite{kaur2022idecode} proposed a novel, base Non-Conformity Measure (NCM) score for detecting the OOD nature of an input as the error in the ID equivariance learned by a model with respect to a set of transforms. Each transform yields an NCM score which is used to create an NCM vector. All the NCM scores are aggregated to compute a final NCM, which is used to decide whether a given test point is OOD or not. \\ \vspace{-2mm} 

\noindent\textbf{iv. Other} methods encompass all the remaining work focusing on adversarial and OOD inputs together~\cite{ratzlaff2019hypergan,sensoy2020uncertainty}. They include methods such as parameter modelling of DNNs~\cite{ratzlaff2019hypergan}, as well as innovative DNN models~\cite{sensoy2020uncertainty}. \\ \vspace{-2mm}

Based on this taxonomy, we structure the remaining sections as follows. Section~\ref{sec:experimental_details} introduces the datasets, models, adversarial attack methods, and evaluation metrics commonly utilised in experiments across existing literature on robust OOD detection and unified robustness. Following that, we provide a detailed analysis of the existing work on robust OOD detection in Section~\ref{sec:robust_ood_detection} and unified robustness in Section~\ref{sec:unified_robustness}. In Section~\ref{sec:limitations_future}, we discuss the limitations of current approaches and emphasise areas requiring more focus, along with novel directions for future research.

\section{Experimental setups}
\label{sec:experimental_details}
\subsection{Datasets, Models, and Adversarial attacks}

First, we discuss the datasets, deep learning models, and adversarial input generation methods that have been commonly used in the work we survey to evaluate the effectiveness of robust OOD and unified robust approaches.

\subsubsection{\textbf{Datasets}}

In OOD work, standard ML datasets are used in different settings. For instance, CIFAR-10~\cite{cifar10}, CIFAR-100~\cite{cifar100}, and ImageNet~\cite{deng2009imagenet} are commonly employed as in-distribution object datasets. To establish OOD benchmarks for these ID datasets, researchers frequently use datasets such as SVHN~\cite{SVHN}, LSUN~\cite{yu2015lsun}, iSUN~\cite{xu2015turkergaze}, Places365~\cite{zhou2017places}, Textures~\cite{cimpoi2014describing}, and CelebA~\cite{liu2015deep}. In the context of digit classification, prior work has commonly relied on MNIST~\cite{deng2012mnist}, SVHN~\cite{SVHN}, and USPS~\cite{uspsdataset} as ID datasets, while the OOD datasets consist of notMNIST~\cite{bulatov2011notmnist} and Fashion-MNIST~\cite{xiao2017fashion}. 

\subsubsection{\textbf{Base models}}

In terms of the choice of deep learning models for the two tasks, prior research has widely adopted well-established architectures such as DenseNet~\cite{huang2017densely}, ResNet~\cite{he2016deep}, WideResNet~\cite{zagoruyko2016wide} and AlexNet~\cite{krizhevsky2012imagenet}. Additionally, the OpenOOD library, which plays a significant role in benchmarking OOD detection methods, utilises the ResNet18~\cite{he2016deep} variant for datasets such as CIFAR10, CIFAR100, and ImageNet-200~\cite{Le2015TinyIV} (a.k.a., Tiny ImageNet). For more challenging ImageNet-1K, the library employs either ResNet50~\cite{he2016deep} or ViT-B/16 models~\cite{dosovitskiy2020image}.

\subsubsection{\textbf{Adversarial attacks}}
To evaluate robust OOD detection methods and unified robust methods, their effectiveness against various adversarial attacks must be tested. The most commonly used attack in literature is the Projected Gradient Descent (PGD)~\cite{madry2017towards}. 

The PGD attack typically starts at a random point within an $\epsilon$-neighbourhood around an original input data point $x$ and iteratively applies small perturbations proportional to the sign of the gradient of the loss with respect to the input. Following each perturbation step, should the resulting perturbed input exceed the $\epsilon$-neighbourhood, a projection function is employed to bring the perturbed input back into the $\epsilon$-neighbourhood. Through this iterative process of perturbation and projection, as shown in Equation~\ref{eq:pgd}, PGD aims to maximise the classification loss of the adversarial example.
\begin{equation}
\label{eq:pgd}
    \Tilde{x}^{t+1} = \Pi_{x+\epsilon} (\Tilde{x}^t + \alpha \cdot sgn(\nabla_{\Tilde{x}^t} \mathcal{L}(f(\Tilde{x}^t;\theta^{*}), y))
\end{equation} 
Here, $\Tilde{x}^0$ is a random point within the $\epsilon$-neighbourhood of $x$, $\alpha$ is the perturbation step size for a single iteration and $\Pi$ projects back the intermediate output to the $\epsilon$-ball.

In addition to PGD, previous work has also employed other attack methods, including the Fast Gradient Sign Method (FGSM)~\cite{fgsm}, Basic Iterative Method (BIM)~\cite{kurakin2018adversarial}, Momentum Iterative Method (MIM)~\cite{dong2018boosting}, Carlini and Wagner (CW) attack~\cite{carlini2017towards}, DeepFool~\cite{moosavi2016deepfool}, and Zeroth Order Optimisation (ZOO) attack~\cite{chen2017zoo}. Most of these attacks being gradient-based, it is important to note that adversaries employ \textbf{gradients of the OOD score} function with respect to the input rather than the \textbf{gradients of the loss} with respect to the input when crafting adversarial inputs to evade the OOD detector. Furthermore, some studies have adopted more sophisticated attack ensembles such as AutoAttack~\cite{croce2020reliable} to evaluate the adversarial robustness.

\subsection{Evaluation metrics}

Here, we present a set of performance metrics to assess the effectiveness of robust OOD detection and unified robust methods. Many of these metrics are commonly applicable to both tasks, ensuring a consistent and standardised evaluation process. 
In order to align with the conventional definition (i.e., Equations~\ref{eq:ood_detection} and~\ref{eq:robust_ood_detection}), we designate ID and benign inputs as positive, and OOD and adversarial inputs as negative. \\ \vspace{-2mm}

\noindent\textbf{Detection error}. This measures the proportion of misclassified samples, both false positives (i.e., either OOD samples getting incorrectly classified as ID or adversarial inputs getting incorrectly classified as benign) and false negatives (i.e., either ID samples getting incorrectly classified as OOD or benign inputs getting incorrectly classified as adversarial). A lower Detection Error indicates a more accurate detector with better differentiation between positive and negative samples. The false positive rate is commonly reported as the \textit{attack success rate} (ASR) in previous research on adversarial attacks, while the true negative rate is often referred to as the \textit{OOD detection rate} in the literature on OOD detection. 
\\ \vspace{-2mm}

\noindent\textbf{FPR95}. This metric calculates the false positive rate (FPR) when the true positive rate (TPR) is set at 95\%. 
In the context of OOD detection, it denotes the percentage of OOD samples incorrectly classified as ID when 95\% of the ID samples are accurately predicted. A lower FPR at 95\% TPR signifies a more effective detector, indicating fewer false positives while maintaining a high true positive rate. Alternatively, some studies present the true negative rate (TNR) at 95\% TPR, which is the complement of FPR (i.e., TNR = 1 - FPR). In this scenario, a higher TNR at 95\% TPR is desirable as it reflects a greater proportion of correctly classified negative instances at a high sensitivity threshold. Moreover, in robust OOD detection, Chen et al.~\cite{chen2020robust} employ this exact metric and term it as \textbf{1-FPR} (at 95\% TPR) to quantify the rate of misclassifying adversarial ID inputs as OOD samples.  \\ \vspace{-2mm}

\noindent\textbf{AUROC}. Area Under the Receiver Operating Characteristic curve is a threshold-independent metric~\cite{davis2006relationship}. It quantifies the classifier's ability to discriminate between positive and negative samples by plotting the True Positive Rate (TPR) against the False Positive Rate (FPR) at various classification thresholds. The AUROC value ranges from 0 to 1, where a higher value indicates better discriminatory power. An AUROC value of 0.5 represents random guessing, while a value of 1 signifies perfect detection. Additionally, prior work has investigated the worst-case AUROC (WC-AUROC), representing the minimal achievable AUROC when each OOD sample is perturbed to achieve maximum confidence within a specified threat model. In this context, the adversarial AUROC (A-AUROC) acts as the \textit{upper bound}, while the guaranteed AUROC (G-AUROC) serves as the \textit{lower bound} for the worst-case performance.  \\ \vspace{-2mm}

\noindent\textbf{AUPR}. The Area Under the Precision-Recall Curve measures the precision-recall trade-off by plotting precision against recall at various classification thresholds. AUPR provides a more informative evaluation when the positive class is rare compared to the negative class. A higher AUPR value indicates better classifier performance, with higher precision and recall. 
\\ \vspace{-2mm}

In the context of robust OOD detection, AUROC and FPR95 are the two most commonly reported metrics, both measuring how well an OOD detector differentiates between ID and OOD inputs, even when they are adversarially perturbed. Similarly, in the unified robustness setting, two of these metrics are used together to report the performance against adversarial and OOD inputs. For instance, detection error can assess how well a model performs against adversarial inputs, while AUROC can measure the model's performance in detecting OOD inputs.

\section{Robust OOD detection}
\label{sec:robust_ood_detection}

\begin{figure}[t]
\centering
\includegraphics[width=0.57\textwidth]{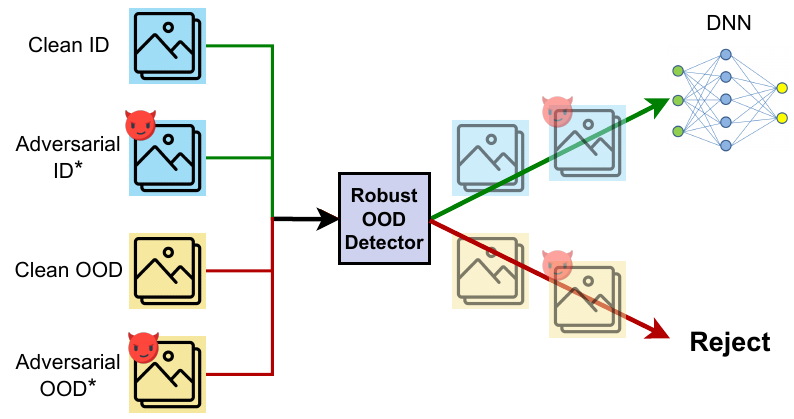} 
\caption{\centering A robust OOD detector correctly detects ID and OOD inputs even if they are adversarially perturbed. \break({\bf Note} - * denotes  adversarial inputs attempting to evade the OOD detector.) } 
\label{fig:robust_ood_detection}
\end{figure}

Robust OOD detection involves accurately distinguishing between ID and OOD inputs, even when subjected to adversarial perturbations ({\bf cf.} Section~\ref{sec:taxonomy}). As illustrated in Figure~\ref{fig:robust_ood_detection}, a robust OOD detector should effectively identify four types of inputs: clean ID, adversarial ID, clean OOD, and adversarial OOD inputs. It is important to emphasise that these adversarial perturbations are designed to evade the OOD detector, unlike conventional adversarial examples that aim to induce misclassifications in the primary classifier. In robust OOD detection, both clean and adversarial ID inputs should be forwarded to the primary classifier, while both clean and adversarial OOD inputs should be discarded. We classify such methods into four branches: outlier exposure-based, learning-based, score-based methods, and an additional class to represent the remaining diverse approaches collectively.

\subsection{Outlier exposure-based methods}
\label{subsec:outlier_exposure_robust}

The core idea in outlier exposure-based methods involves leveraging an auxiliary outlier dataset during model training. These techniques are commonly employed in standard OOD detection~\cite{outlierexposure,zhang2023mixture,yang2021semantically,yu2019unsupervised}. With appropriate modifications, similar methods have been adapted for robust OOD detection. In general, outlier exposure-based methods exhibit superior performance compared to other approaches, as the model gains exposure to OOD data to some degree during training. However, they also possess limitations in generalisation, as the model becomes accustomed to only specific types of OODs.

The work by Chen et al.~\cite{chen2021atom} can be considered one of the key contributions in outlier exposure-based robust OOD detection. Their method, \textbf{A}dversarial \textbf{T}raining with informative \textbf{O}utlier \textbf{M}ining (ATOM) is motivated by the observation that the majority of auxiliary OOD examples may not substantially improve, and in some instances, may even degrade, the decision boundary of the OOD detector. Therefore, there arises a necessity to carefully select outliers (commonly referred to as \textbf{outlier mining}) that provide more insight into learning the OOD decision boundary. Through outlier mining, the authors target selective outliers positioned near the decision boundary between ID and OOD data, thereby enhancing OOD detection. Conversely, in the absence of outlier mining, \textit{easy} outliers are primarily sampled, resulting in a loosely defined decision boundary for the OOD detector. This holds particular significance in robust OOD detection, where the boundary must maintain a sufficient margin from OOD data to prevent adversarial perturbations from crossing it. 

More specifically, in a $K$-class classification setting, this study considers a ($K+1$)-way classifier network $f$, where the ($K+1$)-th class label denotes the \textbf{OOD class} and its prediction confidence is used as the OOD score. During training, an auxiliary dataset $\mathcal{D^{\text{aux}}_{\text{out}}}$ is utilised as outliers, where hard outliers are selectively filtered and incorporated into training for the subsequent epoch. In each training epoch, $N$ data points are randomly sampled from the auxiliary OOD dataset, and the corresponding OOD scores are inferred from the current model. Subsequently, the data points are sorted based on their OOD scores, and a subset of $n$ ($<$$N$) data points is selected, starting from the $q$-th position in the sorted list. These selected samples are then utilised as OOD training data for the subsequent epoch of training. The hyperparameter $q$ plays a crucial role in determining the \textit{informativeness} of the sampled points for the OOD detector; a larger value of $q$ results in less informative sampled examples. Once the outliers are selected, half of them are perturbed using PGD and the network is trained to categorise these perturbed instances as belonging to the ($K+1$)-th class (i.e., \textbf{adversarial training on outliers}). During testing, the prediction confidence scores associated with the ($K+1$)-th class are used as a reliable metric for assessing the model's uncertainty and identifying instances outside the known data distribution, even in the presence of adversarial perturbations. Experimental results show that ATOM outperforms existing methods by a considerable margin across different types of adversarial OOD inputs. For instance, when evaluated on the CIFAR-10 ID dataset, ATOM achieves a reduction of 53\% in FPR95 under $l_{\infty}$ attacks on OOD inputs. Similarly, for CIFAR-100, this reduction is approximately 37\%.

\begin{figure}[t]
\centering
\includegraphics[width=0.59\textwidth]{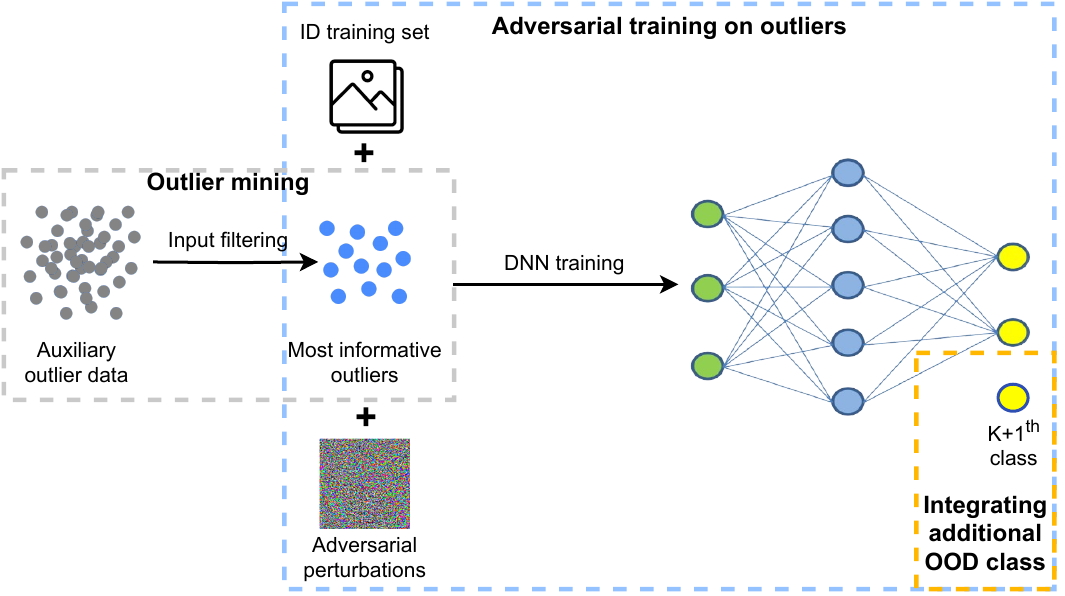} 
\caption{\centering Key concepts utilised in ATOM~\cite{chen2021atom} for robust OOD detection: i. Outlier mining ii. Adversarial training on outliers iii. Integrating an additional OOD class.} 
\label{fig:outlier_exposure}
\end{figure}

This work employs several foundational concepts frequently encountered in the literature on robust OOD detection, as shown in Figure~\ref{fig:outlier_exposure}. Specifically, it explores \textit{outlier mining}, \textit{the integration of an additional OOD class} into the classifier, and \textit{adversarial training on outliers} as foundation strategies. Others have explored various combinations of these strategies in their attempts to address the challenge of robust OOD detection~\cite{hein2019relu,augustin2020adversarial}. For example, Hein et al.~\cite{hein2019relu} provide an analysis of why ReLU neural networks tend to produce high confidence scores for OOD samples and propose Adversarial Confidence Enhancing Training (ACET), which performs adversarial training on the OOD training set to mitigate this issue and identify adversarially perturbed OOD samples. More recent work~\cite{augustin2020adversarial,chen2020robust,yin2022learning,megyeri2021robust,sehwag2019analyzing} further integrates adversarial training into both the ID and OOD training sets to enhance the interpretability and robustness of confidence estimates in DNNs. Unlike ATOM, which incorporates a ($K+1$)-th class, the aforementioned methods use the uniform distribution as the ground truth for OOD data. Together, these methods ensure the robustness of OOD detection through adversarial training on the OOD training set. 

Furthermore, Dionelis et al.~\cite{dionelis2022frob} propose the Few-shot ROBust (FROB), which follows an outlier mining approach for robust OOD detection. FROB utilises a discriminative model for both ID classification and OOD detection, along with a generator responsible for generating OOD samples and learning the support (i.e., the region containing all the training data) boundary for ID classes. Specifically, this generator is trained to generate low-confidence adversarial OOD samples on the support boundary. Instead of relying on a large outlier dataset, which involves an arbitrary selection of outliers, FROB adopts learned negative data augmentation and self-supervised learning to model the complement of the support of ID data. This approach corresponds to outlier mining, aimed at identifying more informative outliers from a large outlier dataset.

The methods discussed so far contribute to enhancing the \textit{empirical OOD robustness} of DNNs, relying on practical observations and experiments, often through trial and error. However, they lack formal guarantees, highlighting the need for \textit{certifiably robust} methods against adversarial OOD inputs that provide formal assurances of robustness for DNNs. Certifiably robust methods~\cite{bitterwolf2020certifiably,meinke2022provably} are designed to offer assurances that the model will perform accurately under specified conditions, irrespective of the attacker's capabilities or knowledge. Notably, the work by Bitterwolf et al.~\cite{bitterwolf2020certifiably} stands out for guaranteeing the adversarial robustness of OOD data within an $l_{\infty}$ ball around such data. The authors introduce a method to achieve guaranteed OOD detection (GOOD) in a worst-case setting. For this purpose, they utilise interval bound propagation (IBP)~\cite{gowal2018effectiveness} to derive a provable upper bound on the maximal confidence of the classifier within an $l_{\infty}$ ball of radius $\epsilon$ centered around a given data point. Interval bound propagation offers entry-wise lower and upper bounds for the output of the $k^{\text{th}}$ layer of a neural network, considering variations of the input $x$ within the $l_{\infty}$ ball of radius $\epsilon$. Similarly, Meinke et al.~\cite{meinke2022provably} introduce Provable OOD Detector (ProoD), which combines a certified binary discriminator for distinguishing between ID and OOD samples with a standard DNN for ID classification. Here, the binary discriminator is trained using IBP to achieve certifiable adversarial robustness in OOD detection. 

Overall, outlier exposure-based methods in robust OOD detection exhibit similar advantages and drawbacks as they do in standard OOD detection. For instance, by training models on datasets containing outliers alongside ID data, DNNs learn to distinguish between usual patterns and unexpected anomalies. However, in the presence of adversarial perturbations, this distinguishability may not be as evident as in standard OOD detection. Moreover, these methods are relatively straightforward to implement, often integrated with existing DNNs through fine-tuning with the outlier data. A prominent example is performing adversarial training on outliers. In adversarial training, a well-known trade-off exists between generalisation (i.e., test accuracy) and robustness (i.e., adversarial accuracy)~\cite{madry2017towards}. Even with adversarial training using only ID data, enhancing adversarial robustness often necessitates sacrificing a considerable amount of test accuracy. Therefore, incorporating OOD data into adversarial training may further diminish the DNNs' generalisability on ID data.

Additionally, the effectiveness of these methods heavily relies on the quality of the auxiliary outlier data. The chosen outliers need to be representative of potential real-world OOD examples, and the dataset itself should be comprehensive enough. However, curating such a dataset is highly challenging, and the auxiliary dataset is less likely to cover the entire OOD space.

\subsection{Learning-based methods}
\label{subsec:learning_based_robust}

Unlike outlier exposure-based methods, learning-based approaches do not rely on auxiliary outlier knowledge during training to ensure robust OOD detection. Instead, they employ alternative strategies to train the primary classifier on ID data, diverging from the conventional supervised cross-entropy-based learning.

For instance, Khalid et al.~\cite{khalid2022rodd} demonstrated that \textit{self-supervised adversarial contrastive learning} can improve OOD detection robustness. The authors first train an encoder using a contrastive loss function on ID data. This training maximises the similarity between the low dimensional representations of augmented and PGD-perturbed ID data. Unlike standard adversarial training, which minimises the classification loss for perturbed inputs, this approach attempts to push covariate-shifted (i.e., augmented or perturbed) ID data closer in the learned representation space. The authors then use the concept of the \textit{union of one-dimensional embeddings} to project the deep features of different classes from the encoder onto \textbf{one-dimensional}, \textbf{mutually orthogonal} predefined vectors representing each class. The purpose of this uni-dimensional mapping is to ensure that the intra-class distribution consists of samples that align most closely with the one-dimensional vector characterising their class, and the orthogonality ensures that there are wide angles between one-dimensional embeddings of different classes resulting in a large rejection region for OOD samples within the inter-class space. To achieve this, the penultimate layer of the model is modified using cosine similarity and a sharpening layer which effectively maps the encoder output to a scalar value. 

After the training stage, the authors employ singular value decomposition (SVD) to compute the first singular vector of each class using the encoder outputs of training data. OOD testing is then performed by assessing the inner products between these calculated first singular vectors, which represent their respective classes, and the extracted encoder features for the test sample. This inner product is compared against a predefined threshold to determine whether a given test input is OOD or not. The improvement in OOD robustness stems from the employed self-supervised contrastive learning approach. The experimental findings demonstrate that the proposed approach achieves around a 25\% reduction in FPR95 in robust OOD detection compared to a baseline in standard OOD detection. However, the authors only assess the robustness against ID and OOD data corrupted with common types of distortions, such as Gaussian noise, motion blur, and snow. Notably, they do not consider adversarial perturbations, which are often regarded as indicative of the worst-case performance~\cite{hendrycks2019benchmarking}. Consequently, their evaluation captures only an average performance rather than the worst-case performance of robust OOD detection.

Moreover, Berrada et al.~\cite{berrada2021make} introduce a framework designed to handle probabilistic networks like Bayesian neural networks (BNNs) and Monte Carlo (MC) Dropout networks, as well as uncertain inputs characterised by distributional shifts due to noise or other perturbations. For a trained, verification-agnostic (i.e., not designed to be verifiable) probabilistic neural network, this study proposes a novel verification technique that extends Lagrangian duality, replacing standard Lagrangian multipliers with \textit{functional multipliers} that can be arbitrary functions of the activations at a given layer. For the verification, the authors employ a layer decomposition of the network such that intermediate inner optimisation problems can be solved in closed form using linear Lagrangian multipliers. For OOD detection, they leverage thresholding on the softmax value maximised across labels, enabling the classification of samples as OOD even under PGD-based adversarial perturbations within an $l_{\infty}$ norm ball. The effectiveness of their approach is validated empirically on BNNs and MC-Dropout networks, where they certify properties such as robust OOD detection. The results significantly surpass previous works, such as enhancing the certified AUROC for robust OOD detection by approximately 30\% for a VGG-64 MC-Dropout CNN trained on CIFAR-10. Similarly, they achieve notable improvement in the $l_{\infty}$ robust accuracy, increasing it by approximately 15\% for a BNN trained on MNIST. Given that this method entails making adjustments to the layers of pre-trained networks, we categorise it within learning-based robust OOD detection methods.

\subsection{Score-based methods}
\label{subsec:score-based_robsut}

In Section~\ref{subsec:outlier_exposure_robust}, we covered methods that leverage knowledge from outlier data during training, followed by approaches in Section~\ref{subsec:learning_based_robust} that do not rely on outlier data but instead introduce modifications to the conventional training process for robust OOD detection. Next, we describe robust OOD detection techniques that can complement conventionally trained models (i.e., those trained through supervised learning using standard cross-entropy). These techniques, known as post-hoc methods, primarily involve generating a score using information extracted from the trained model, such as gradients, logits, or intermediate outputs, to perform robust OOD detection.

For instance, Fort et al.~\cite{fort2022adversarial} showed that some of the existing score-based OOD baselines such as maximum softmax probability (MSP)~\cite{hendrycks2016baseline}, Mahalanobis distance-based score (MDS)~\cite{lee2018simple} which perform well in detecting clean OOD inputs, drastically fail under adversarial perturbations. This study shows that even simple attacks such as FGSM~\cite{fgsm} can drop the AUROC performance of MDS method by $\approx$57\%. As a solution, the authors show that \textit{relative Mahalanobis distance-based score} (RMDS)~\cite{ren2021simple} can be used as a robust OOD detector to handle OOD inputs even in the presence of adversarial perturbations. 

More specifically, let $z$ be the low dimensional representation from a neural network's penultimate layer for a given test input $x$. Then RMDS is defined as,
\begin{equation}
    RMDS_k(z) = MDS_k(z) - MDS_0(z)
\end{equation}
where $MDS_k$ represents the Mahalanobis distance to the class-specific Gaussian distribution of class $k$ and $MDS_0$ indicates the Mahalanobis distance to a Gaussian distribution fitted to the entire in-distribution dataset. It is important to note that this study uses the \textit{gradients of the OOD score with respect to the input} to conduct FGSM attacks instead of the gradients of the loss with respect to the input. The authors further report that regardless of how expressive the embeddings generated by powerful networks such as Vision Transformers (ViT) are, they are still susceptible to adversarial examples, particularly generated using near-OOD inputs. Finally, the authors empirically show that working with lower resolution images and OOD score ensembles (i.e., Average of OOD scores from two well-performing models) increases the adversarial robustness of OOD detection.

Furthermore, Azizmalayeri et al.~\cite{azizmalayeri2022your} present a secondary OOD detector known as the Adversarially Trained Discriminator (ATD), which utilises the MSP score for robust OOD detection. ATD leverages the \textit{pre-trained primary model to extract robust features} and a generator model to produce OOD samples. This approach is motivated by the notion that an ideal defence should expose the model to a wide range of adversarial perturbations based on both ID and OOD samples during training. To achieve this, ATD's training incorporates three types of inputs: real ID inputs, real outliers, and generated outliers. The authors emphasise that, for training stability, the adversarial training of the discriminator should attack both real ID and real outliers but not generated outliers. Consequently, ATD achieves robustness from two perspectives: implicit robustness for various OOD inputs via the generator and robustness through adversarial training using perturbed ID and OOD data. 

When applied with CIFAR-10 and CIFAR-100 as ID data, ATD significantly outperforms all previous methods in robust AUROC while maintaining high standard AUROC and classification accuracy. As ATD serves as a \textbf{secondary} classifier for robust OOD detection, we refrain from categorising it under outlier exposure or learning-based methods, as both those categories relate to the training of the primary classifier. However, given that robust OOD detection relies on the MSP of the ATD, we classify it under score-based approaches.

\subsection{Other}
\label{subsec:other_robust}

These methods involve developing  novel adversarial attacks aimed at circumventing existing OOD detectors both in digital~\cite{yoon2022evaluating,kaya2022generating} and physical domains~\cite{ojaswee2023benchmarking}

For example, Kaya et al.~\cite{kaya2022generating} propose an adaptive attack termed the \textit{statistical indistinguishability attack} (SIA), which aims to minimise the statistical distance between the distributions of adversarial and natural samples. Notably, SIA simultaneously targets all DNN layers, as this study demonstrates that adversarial examples indistinguishable at one layer may fail to be so at others. Therefore, SIA optimises DNN representations for adversarial examples at each layer to resemble those of clean inputs. Empirical results show SIA's effectiveness in evading four individual adversarial detectors, two dataset shift detectors, and an OOD detector.

Similarly, Yoon et al. \cite{yoon2022evaluating} propose a generative approach called \textit{Evaluation-via-Generation} (EvG) to assess the robustness of an OOD detector. EvG employs a variation model capable of performing affine transformations, colour jittering, or generating new data points by combining existing OOD data using a Generative Adversarial Network (GAN). This variation model captures a distribution with plausible adversarial variations of a test OOD dataset. Subsequently, a Markov Chain Monte Carlo (MCMC) optimisation algorithm is utilised to draw a sample from this distribution with the lowest detector score, representing the most challenging OOD sample for the detector. AUROC scores derived from evaluating the OOD detectors with samples obtained from this process reveal that some robust OOD detectors (i.e., showing AUROC $>$ 0.9) perform poorly compared to their performance on clean data. Additionally, the authors employ the rank of the smallest detection score (i.e., derived from the worst-performing OOD sample, which is considered the closest to ID) to evaluate OOD detectors. 

In contrast to the above digital domain attacks, Ojaswee et al.~\cite{ojaswee2023benchmarking} propose a novel benchmarking dataset consisting of OOD inputs perturbed using physical adversarial patches of different variations. These adversarial patches are crafted by modifying the styles of everyday object images like cups, plates, and socks. The inclusion of diverse style patches ensures that defence algorithms are not biased toward a single type of adversarial patch. These patches are then carefully integrated into clean inputs to execute physical adversarial attacks. Accordingly, the authors create a comprehensive dataset comprising 44K images, including both real and adversarially patched examples. Empirical evaluations demonstrate that OOD detectors exhibit significantly reduced performance in detecting unseen adversarial patches during model training.

\subsection{Summary of Robust OOD Detection}

In this section, we first discussed {\bf outlier exposure-based methods}, which operate at the data level prior to model training. These methods have the unique advantage of utilising the additional knowledge from outlier training data. Secondly, we investigated {\bf learning-based techniques}, which operate at the model level during model training. These methods aim to model the support of ID data using sophisticated training strategies without relying on additional outlier knowledge. Both outlier exposure-based and learning-based methods have the ability to adjust decision boundaries to accommodate robust OOD detection, with the former exhibiting greater generalisability than the latter. Thirdly, we explored post-process score-based techniques designed to perform robust OOD detection using a pre-trained model not specifically trained for that purpose. Consequently, these methods may exhibit lower performance due to inherent limitations in the model's ability to distinguish between ID and OOD data. However, they are the easiest to implement with any pre-trained models without any specific prerequisites. Finally, we discussed evaluation strategies in both digital and physical domains aimed at ensuring the adversarial robustness of existing OOD detectors. Table~\ref{tab:rob_ood_summary} provides an overview of the methods for robust OOD detection.

\begin{table*}[t!]
\scriptsize
  \centering
  \caption{\centering Summary of robust OOD detection methods. \break Note that OOD datasets and performance are reported for the ID dataset marked in \textbf{bold}.}
  \label{tab:rob_ood_summary}
  \resizebox{\textwidth}{!}{%
    \begin{tabular}{p{3cm}|p{6.5cm}|p{2cm}|p{2.2cm}|p{2.0cm}|p{2.5cm}}
      \toprule
      \textbf{Research Work} & \textbf{Summary} & \textbf{ID datasets} & \textbf{OOD datasets} & \textbf{Adversary} & \textbf{Performance} \\
      \midrule
      \multicolumn{6}{c}{\textbf{Outlier exposure-based}} \\
      \midrule
      \midrule
      Chen et al.~\cite{chen2021atom} & Introduce outlier mining to extract the most informative outliers to perform adversarial training. An additional OOD class is incorporated as the ground truth for OOD inputs. & \textbf{CIFAR-10}, CIFAR-100 & SVHN, Textures, Places365, LSUN (crop \& resize), and iSUN & $l_{\infty}$-PGD \break $\epsilon = 8/255$ & FPR95 - 20.55\% \break AUROC - 88.94\% \\
      \midrule
      Hein et al.~\cite{hein2019relu} & Perform adversarial training only on OOD inputs. Uniform distribution is considered as the ground truth for OOD inputs. & \textbf{CIFAR-10}, CIFAR-100 & SVHN, Textures, Places365, LSUN (crop \& resize), and iSUN & $l_{\infty}$-PGD \break $\epsilon = 8/255$ & FPR95 - 74.45\% \break AUROC - 78.05\%  \\
      \midrule
        Chen et al.~\cite{chen2020robust} & Perform adversarial training on both ID and OOD inputs. Uniform distribution is considered as the ground truth for OOD inputs. & \textbf{CIFAR-10}, CIFAR-100, GTSRB & SVHN, Textures, Places365, LSUN (crop \& resize), and iSUN. & $l_{\infty}$-PGD \break $\epsilon = 1/255$ & FPR95 - 41.59\% \break AUROC - 92.69\% \\
        \midrule
        Augustin et al.~\cite{augustin2020adversarial} & Perform adversarial training on both ID and OOD data against an $l_2$ adversary. Uniform distribution is considered as the ground truth for OOD inputs. & \textbf{CIFAR-10}, SVHN, CIFAR-100, ImageNet & SVHN, LSUN (crop), CIFAR-100, ImageNet & $l_{2}$-PGD \break $\epsilon = 1.0$ & AUROC - 84.30\% \\
        \midrule
        Yin et al.~\cite{yin2022learning} & Demonstrate binary adversarial training's capability to learn an energy function that models the support of ID data, enhancing OOD adversarial robustness. & \textbf{CIFAR-10} & SVHN, CIFAR-100, ImageNet & $l_{2}$-AutoAttack\break $\epsilon = 1.0$ & AUROC - 81.53\% \\
        \midrule
        Megyeri et al.~\cite{megyeri2021robust} & Show that the key factor for adversarially robust OOD detection is not the OOD training or detection method itself but rather the application of matching detection and training methods. & \textbf{CIFAR-10}, MNIST & SVHN & $l_{\infty}$-PGD \break $\epsilon = 8/255$ & AUROC - 86.00\% \\
        \midrule
        Sehwag et al.~\cite{sehwag2019analyzing} & Utilise a hybrid approach involving iterative adversarial training to achieve robust classification of both unmodified and adversarial OOD data, particularly towards a background class. & \textbf{CIFAR-10} & MNIST, ImageNet, VOC12, Internet Photographs & $l_{\infty}$-PGD \break $\epsilon = 8/255$ & ASR - 7.40\% \\
        \midrule
        Dionelis et al.~\cite{dionelis2022frob} & Generate the support boundary of the ID class distribution in a self-supervised manner and combine it with few-shot outlier exposure, imposing low confidence at this learned boundary. & \textbf{CIFAR-10}, SVHN & SVHN, CIFAR-100 & $l_{\infty}$-PGD \break $\epsilon = 0.01$ & A-AUROC - 86.00\%\break G-AUROC - 71.80\% \\
        \midrule
        Bitterwolf et al.~\cite{bitterwolf2020certifiably} & Apply interval bound propagation (IBP) to ensure certifiable worst-case guarantees for OOD detection by ensuring low confidence not only at the OOD point but also within an $l_{\infty}$-ball around it. & \textbf{CIFAR-10}, SVHN, MNIST & SVHN, CIFAR-100, LSUN (classroom) & $l_{\infty}$-PGD \break $\epsilon = 0.01$ & A-AUROC - 68.58\%\break G-AUROC - 67.78\% \\
        \midrule
        Meinke et al.~\cite{meinke2022provably} & Merge a certified binary OOD detector trained with IBP with the primary classifier for ID classification. This joint classifier achieves provably adversarially robust OOD detection and high ID accuracy. & \textbf{CIFAR-10}, CIFAR-100, ImageNet & SVHN, CIFAR-100, LSUN (classroom) & $l_{\infty}$-PGD \break $\epsilon = 0.01$ & A-AUROC - 49.85\%\break G-AUROC - 48.98\% \\
    \midrule
    \multicolumn{6}{c}{\textbf{Learning-based}} \\
    \midrule                 
    \midrule
      Khalid et al.~\cite{khalid2022rodd} & Demonstrate that a pre-trained model using self-supervised adversarial contrastive learning yields superior low-dimensional feature representations, thereby enhancing OOD robustness. & CIFAR-10, \textbf{CIFAR-100} &  TinyImageNet (crop \& resize), LSUN (resize), Places, Textures, SVHN, and iSUN & Perturbations such as Gaussian noise, motion blur, and snow & FPR95 - 57.49\% \break AUROC - 81.60\% \\
      \midrule
      Berrada et al.~\cite{berrada2021make} & Propose a verification technique that extends Lagrangian duality, replacing standard Lagrangian multipliers with functional multipliers that can be arbitrary functions of the activations at a given layer.  & \textbf{CIFAR-10} & CIFAR-100 & $l_{\infty}$-PGD \break $\epsilon = 0.001$ & A-AUROC - 26.97\%\break G-AUROC - 48.98\% \\
    \midrule
    \multicolumn{6}{c}{\textbf{Score-based}} \\
      \midrule
    \midrule
      Fort et al.~\cite{fort2022adversarial} & Show the adversarial vulnerability of current OOD detectors and emphasise that relative Mahalanobis distance and OOD detector ensembles exhibit better adversarial robustness in OOD detection. & \textbf{CIFAR-100} & CIFAR-10 & $l_{\infty}$-FGSM \break $\epsilon = 1/255$ & AUROC - 71.84\% \\
      \midrule
      Azizmalayeri et al.~\cite{azizmalayeri2022your} & Introduce the Adversarially Trained Discriminator (ATD) for robust OOD detection, employing a pre-trained robust model to extract features and a generator model to generate OOD samples. & \textbf{CIFAR-10}, CIFAR-100 & MNIST, LSUN TinyImageNet, Places365, iSUN, Birds, Flowers & $l_{\infty}$-PGD \break $\epsilon = 8/255$ & AUROC - 69.30\% \\
    \midrule
    \multicolumn{6}{c}{\textbf{Other}} \\
      \midrule
    \midrule
        Kaya et al.~\cite{kaya2022generating} & Propose an adaptive attack named SIA against OOD detectors to create adversarial examples that follow the same distribution as the ID inputs with respect to DNN representations.& \textbf{CIFAR-10} & SVHN & $l_{\infty}$-SIA \break $\epsilon = 0.01$ & Energy detector \break FPR95 - 95.00\% \\
        \midrule
        Yoon et al.~\cite{yoon2022evaluating} & Propose Evaluation-via-Generation to assess OOD detector robustness. EvG employs a generative model for outlier synthesis and MCMC sampling to identify misclassified outliers with high confidence. & \textbf{CIFAR-10} & SVHN, CelebA & Outliers created using EvG & ATOM detector \break AUROC - 69.35\% \\
        \midrule
        Ojaswee et al.~\cite{ojaswee2023benchmarking} & Propose a novel dataset using adversarial patches of different variations in the physical world to evaluate the DNN robustness against stealthy OOD images. & \textbf{ImageNet} & COCO &  Adversarially patched COCO data samples & Detection rate - 69.59\% \\
      \bottomrule
    \end{tabular}%
  }
\end{table*}

\begin{center}
\begin{boxM}
\footnotesize
\centering \textbf{\underline{Robust OOD detection}} 
\vspace{8pt}
\begin{itemize}
    \item \textbf{Outlier exposure-based} methods utilise outlier training data prior to model training, enhancing robust OOD detection with additional knowledge.
    \begin{itemize}
        \item \textbf{Pros}: Better generalisability.
        \item \textbf{Cons}: Prone to over-fitting to seen outliers, May degrade performance on ID classification.
    \end{itemize}
    \item \textbf{Learning-based} methods aim to model ID data support using complex training strategies. 
    \begin{itemize}
        \item \textbf{Pros}: Do not need to rely on the quality of training outlier data.
        \item \textbf{Cons}: Increased training complexity.
    \end{itemize}
    \item  \textbf{Score-based} post-processing techniques leverage information from pre-trained models to perform robust OOD detection.
    \begin{itemize}
        \item \textbf{Pros}: Easy implementation, High performance on ID classification.
        \item \textbf{Cons}: Relatively low robustness, May require complex post-processing leading to increased latency. 
    \end{itemize}
    \item \textbf{Other} evaluation strategies ensure adversarial robustness across digital and physical domains for existing OOD detectors.
    \begin{itemize}
        \item \textbf{Pros}: Demonstrate the adversarial vulnerability of current OOD detectors, Can be used as a benchmark.
        \item \textbf{Cons}: NA - No counter-measures proposed.
    \end{itemize}
\end{itemize}
\end{boxM}
\end{center}

\section{Unified robustness}
\label{sec:unified_robustness}

\begin{figure}[t]
\centering
\includegraphics[width=0.59\textwidth]{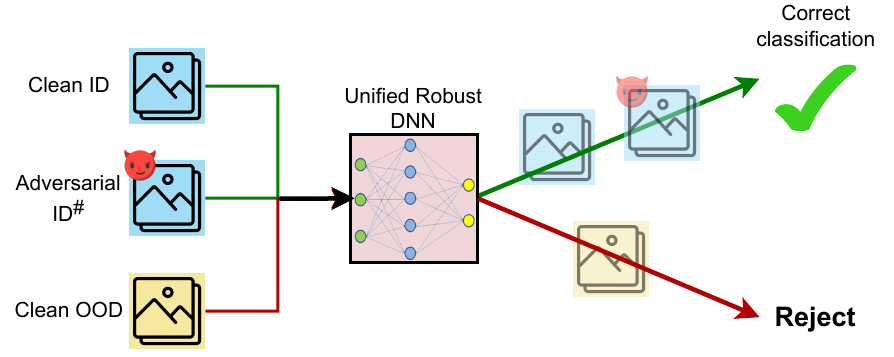} 
\caption{\centering A unified robust method correctly classifies ID inputs even if adversarially perturbed. Furthermore, it identifies and rejects clean OOD inputs. ({\bf Note} - \# denotes adversarial examples targeting the primary classifier.)}
\label{fig:unified_robustness}
\end{figure}

In Section~\ref{sec:robust_ood_detection}, we explored methods capable of effectively detecting both clean ID and OOD inputs, as well as adversarial inputs attempting to bypass the OOD detector (i.e., robust OOD detection). In contrast, within unified robustness, we investigate techniques that simultaneously make the models robust against \textbf{clean OOD and adversarial ID} inputs, as shown in Figure~\ref{fig:unified_robustness}. Unlike in robust OOD detection, these adversarial ID inputs are crafted to deceive the primary classifier. The overall objective of unified robustness is to ensure accurate predictions on both clean and adversarial ID inputs while detecting and rejecting clean OOD inputs. It is important to note that these methods do not focus on adversarial OOD inputs. Following a structure similar to that outlined in Section~\ref{sec:robust_ood_detection}, we categorise unified robust approaches into four categories: data augmentation-based, learning-based, score-based, and other various methods.

\subsection{Data augmentation-based methods}
\label{subsec:data_aug_unified}

These methods are employed during the data pre-processing stage, where the training set undergoes augmentation. This additional data serves to improve the model's resilience against distributional shifts. Notably, the \textit{RegMixup} method proposed by Pinto et al.~\cite{pinto2022using} stands out as a key contribution to data augmentation-based techniques. The core principle of RegMixup involves generating interpolated samples from pairs of inputs within the training set. A regularisation term is introduced to the standard cross-entropy loss function, facilitating the model's training on these interpolated samples. Each new sample is created as a linear combination of two training inputs, with the label also being determined as the linear combination of these inputs. 

RegMixup is built upon improving the method named \textit{Mixup}~\cite{zhang2017mixup}, which uses \textit{vicinal risk minimisation}~\cite{chapelle2000vicinal} for training neural networks. Specifically, vicinal risk minimisation (VRM) aims to model a more comprehensive distribution around each input-output pair, enabling a more precise estimation of risk computed within the vicinity of each training sample. In Mixup, for a given input-output pair (i.e., ($x_i$,$y_i$) and ($x_j$,$y_j$)), an interpolated sample ($\Bar{x}$) and its corresponding label ($\Bar{y}$) are generated as,
\begin{align}
\label{eq:mixup}
\begin{aligned}
    &\Bar{x} = \lambda x_i + (1-\lambda)x_j \\
    & \Bar{y} = \lambda y_i + (1-\lambda)y_j  
\end{aligned}
\end{align}
where $\lambda \sim \beta (\alpha,\alpha) \in [0,1]$ is the interpolation factor and $\alpha \in (0,\infty)$. Accordingly, Mixup slightly perturbs a clean sample towards the direction of another sample and trains the model on these interpolated images to improve generalisation. This contrasts with conventional adversarial training, which perturbs the input towards maximising the classification loss. Distinguishing between Mixup and RegMixup, the former solely trains the model on interpolated images, whereas the latter incorporates interpolated images as an additional regulariser alongside standard cross-entropy-based training on original inputs. This modification not only alleviates Mixup's limitation on detecting OOD samples but also significantly improves its performance on various forms of covariate shifts in ID samples. 

Nonetheless, the evaluation of this work does not include performance against adversarial perturbations, which quantifies the worst-case robust performance on ID data. Experimental results in OOD detection demonstrate an improvement of over 9\% in AUROC, compared to Mixup when SVHN is considered as OOD for WideResNet models trained on either CIFAR-10 or CIFAR-100. Moreover, RegMixup achieves a 2.45\% reduction in detection error compared to Mixup against corrupted CIFAR-100 inputs and nearly 1.45\% reduction against corrupted CIFAR-10 inputs.

In contrast to RegMixup, where data augmentation involves mixing two training inputs, Hendrycks et al.~\cite{hendrycks2022pixmix} introduce a data augmentation technique named \textit{PixMix} by blending training data with structurally complex objects such as \textit{fractals} and \textit{feature visualisations}. Fractals are infinite, complex patterns that exhibit self-similarity across varying scales. Additionally, this study utilises feature visualisations derived from the initial layers of various convolutional architectures through OpenAI Microscope.\footnote{https://openai.com/research/microscope} These augmented images are incorporated into model training, resulting in a model that demonstrates near Pareto-optimal performance against safety threats such as adversarial attacks and OOD inputs. For example, on CIFAR-100, PixMix reduces the detection error for $l_{\infty}$ PGD adversarial attacks by 3.9\% compared to a standard data augmentation baseline, where the authors employ random cropping with zero padding followed by random horizontal flips. Regarding OOD detection, PixMix increases the AUROC by 11.6\% compared to the same baseline when CIFAR-100 is considered as ID.

\subsubsection{\textbf{Data reduction}}

Both RegMixup and PixMix improve the robustness against adversarial examples and OOD inputs by augmenting the training dataset. In contrast, Azizmalayeri et al.~\cite{azizmalayeri2023data} propose a novel perspective of \textit{detecting and removing hard samples} from the adversarial training procedure rather than employing complex algorithms to mitigate their effects. During adversarial training, we frequently encounter samples that are close to the decision boundaries or even outside of the training class distributions (i.e., OOD data). Due to the difficulty of training a model on these samples, the authors refer to them as hard training samples and argue that it may be preferable to avoid training on these samples altogether. To identify such inputs, this paper utilises the Correct Class Softmax Probability (CCSP) score. After each epoch in the training loop, CCSP scores are computed for all samples in the training dataset, and a subset of samples with the lowest CCSP scores is excluded from the training dataset only for the subsequent epoch. After model training, adversarial examples and OOD detection are performed using the maximum softmax probability. This method marginally improves the performance of standard PGD-adversarial training. 
Additionally, the authors demonstrate that when the training dataset is augmented with OOD inputs, this method detects and removes 80\% of them from the training set.

\subsection{Learning-based methods}
\label{subsec:learning_based_unified}

In the previous section, we discussed methods that operate at the data level by augmenting the training dataset. In contrast, learning-based methods utilise the original training data and modify the conventional training process of DNNs to improve robustness against various distributional shifts. For instance, Hendrycks et al.~\cite{hendrycks2019using_self_supervised} demonstrate that self-supervised learning can enhance model performance against uncertain inputs such as adversarial examples and other non-adversarial input corruptions. This study also shows that self-supervision provides substantial benefits in detecting OOD samples, especially those that are challenging and closely related to ID. In fact, its performance surpasses that of fully supervised methods. 

In this work, the authors adopt PGD adversarial training~\cite{madry2017towards} as the base setup in their experiments. PGD adversarial training is widely recognised as the most effective empirical defence against adversarial examples to date. In their approach, the primary classification network (i.e., $f_{\text{classifier}}$) is trained alongside a separate auxiliary rotation prediction head (i.e., $f_{\text{rot\_head}}$). This auxiliary head takes the penultimate vector from the network as input and produces a softmax distribution with four categories. It is trained concurrently with the rest of the network to predict the degree of rotation ($r$) (i.e., $r \in \{0^\circ, 90^\circ, 180^\circ, 270^\circ\}$) applied to a PGD-perturbed input image. Consequently, the total loss during training consists of both a supervised loss and a self-supervised loss, represented by the first and second terms, respectively, in Equation~\ref{eq:ss_rotation}.

\begin{equation}
\begin{split}
    \label{eq:ss_rotation}
    \mathcal{L}(f(x;\theta),y) = \mathcal{L}_{\text{CE}}(f_{\text{classifier}}(x'),y) +  \lambda \{\frac{1}{4} \cdot 
    \sum_{r \in \{0^\circ, 90^\circ, 180^\circ, 270^\circ\}} 
    \mathcal{L}_{\text{CE}}(f_{\text{rot\_head}}(R_r(x')),r)\}
\end{split}
\end{equation}

Here, $x'$ denotes a PGD-perturbed instance of an input $x$ whose ground truth is $y$, while $R_r(x)$ represents the operation of rotating $x$ by $r$ degrees. The second term of the equation signifies that each perturbed input undergoes four levels of rotation, and the resulting mean cross-entropy loss for rotation prediction is utilised as the self-supervised loss. To adapt the PGD adversary for the new training setup, the authors modify the loss function utilised in the PGD attack to maximise both the rotation loss and the classification loss. Empirical verification conducted by the authors demonstrates that integrating adversarial training with self-supervised learning reduces the detection error against PGD adversarial examples by 5.6\% compared to standard PGD adversarial training. This approach is similar to adversarial training, incorporating self-supervised learning for regular adversarial robustness~\cite{kim2020adversarial,naseer2020self,chen2020self}. However, these studies did not evaluate their OOD detection capabilities.

For OOD detection, the authors formulate an OOD score ($\mathcal{S}$) combining the predictions from the main classifier and the rotational head as shown in Equation~\ref{eq:ss_rot_ood}.
\begin{equation}
\begin{split}
    \label{eq:ss_rot_ood}
    \mathcal{S}(x) = \text{KL}(U || f_{\text{classifier}}(x)) +  \frac{1}{4} \cdot
    \sum_{r \in \{0^\circ, 90^\circ, 180^\circ, 270^\circ\}} 
    \mathcal{L}_{\text{CE}}(f_{\text{rot\_head}}(R_r(x),r)
\end{split}
\end{equation}
Here, KL refers to the Kullback-Leibler divergence, and $U$ represents the uniform distribution. The findings from Hendrycks et al.~\cite{outlierexposure} indicate that the KL divergence between $U$ and the predicted distribution yields similar performance to the MSP baseline. Therefore, the authors leverage this idea to formulate their OOD score using KL divergence, as it effectively complements the rotation loss. Experimental results in OOD detection demonstrate a 4.8\% improvement in AUROC with self-supervision compared to the maximum softmax probability baseline. 

Additionally, Hendrycks et al.~\cite{hendrycks2019using_pre-trained} demonstrate that leveraging \textit{pre-training} can enhance the resilience of DNNs against uncertain inputs such as adversarial examples and OOD data. Pre-training refers to the initialisation of model parameters before fine-tuning on a specific task or dataset. This process entails initially training a DNN on a larger or related dataset to learn general features and representations, subsequently leading to enhanced performance when applied to the target task. While pre-training may not directly enhance performance on conventional classification metrics, the authors show that it significantly enhances \textit{model robustness and uncertainty estimates}. Accordingly, they introduce the concept of \textbf{adversarial pre-training} for DNNs. Their findings indicate that opting for targeted adversaries or no adversaries during pre-training does not yield significant robustness enhancements.  However, they observe that a network pre-trained adversarially against an untargeted adversary can surpass the previous state-of-the-art model by approximately 10\% in adversarial robustness. Furthermore, regarding OOD detection, the authors empirically show that pre-trained networks significantly outperform the models trained from scratch in terms of AUROC and AUPR.

Both the previous approaches are applicable to the training of deterministic DNNs. In contrast, Malinin et al.~\cite{malinin2019reverse} propose a method to improve the unified robustness in a type of stochastic neural network known as prior networks. Prior networks~\cite{malinin2018predictive} (PNs) were initially proposed to capture both distributional uncertainty in data (i.e., \textit{aleatoric uncertainty}) and uncertainty in models (i.e., \textit{epistemic uncertainty}). This is achieved by parameterising a Dirichlet prior over output distributions, allowing them to emulate an ensemble of models using a single network. In their primary work, Malinin et al.~\cite{malinin2018predictive} train prior networks using \textit{forward} KL divergence between the model and a target Dirichlet distribution. However, to achieve competitive classification performance, it is essential to incorporate auxiliary losses such as cross-entropy along with forward KL divergence. Nevertheless, training PNs using this criterion poses challenges, particularly on complex datasets with numerous classes. As a solution, in their subsequent work~\cite{malinin2019reverse} the authors propose to utilise \textit{reverse} KL divergence between the model and a target Dirichlet distribution to train PNs. This modification allows for the successful training of PNs for classification tasks with any number of classes, while also improving OOD detection performance. Secondly, leveraging this novel training criterion, the paper explores the application of PNs for detecting adversarial attacks and introduces a generalised form of adversarial training. The experimental results demonstrate that constructing adaptive white-box attacks against PNs trained using the proposed method requires significantly more computational effort compared to networks defended using standard adversarial training or MC-Dropout networks.

Both deterministic and stochastic DNNs offer unique advantages and drawbacks in the context of unified robustness. Specifically, deterministic DNNs provide straightforward mappings from inputs to outputs, making them simpler to train, interpret, and deploy. They are efficient in terms of computational resources and well-suited when precise predictions are required. However, deterministic models may struggle with handling uncertainty in data in the face of adversarial and OOD inputs. In contrast, stochastic DNNs introduce randomness into the model, enabling them to quantify uncertainty in predictions, which is more effective in adversarial and OOD detection. Yet, stochastic models tend to be more computationally intensive, complex to implement, and challenging to interpret due to the added randomness. They may also require additional optimisation and computational resources for training and deployment. Ultimately, the choice between deterministic and stochastic DNNs depends on factors such as the importance of uncertainty estimation, computational resources, and interpretability requirements of the task at hand.

Another approach proposed by Lee et al.~\cite{lee2020adversarial} demonstrates that by utilising a hybrid model based on the joint energy-based model and adversarial training, a unified robust model can be trained. Without making any modifications to the original neural network, the authors achieved this by supplementing the network with a learnt energy-based model with a loss function. The energy-based model learns the underlying distribution of the dataset, which can then be sampled through finite step Langevin dynamics~\cite{nijkamp2020anatomy}. Furthermore, the authors propose a new loss function that includes an energy-based term as well as a robust loss term given by the KL divergence between the model output over the normal and adversarial distribution. However, the authors have only demonstrated the ability of their method over smaller datasets. Whereas the adversarial robustness of their model has been superior to the state-of-the-art methods, the unperturbed accuracy of their model is slightly lower.

\subsection{Score-based methods}
\label{subsec:score_based_unified}

In Section~\ref{subsec:data_aug_unified}, we explored methods involving data augmentation prior to training, followed by approaches outlined in Section~\ref{subsec:learning_based_unified} that introduce modifications to the training process to achieve unified robustness. Now, we investigate unified robust techniques that can complement models trained through supervised learning using standard cross-entropy. Analogous to Section~\ref{subsec:score-based_robsut}, these post-hoc methods primarily entail generating a score using information extracted from the trained model, such as gradients, logits, or intermediate outputs, to handle both adversarial and OOD inputs. Notably, the work by Lee et al.~\cite{lee2018simple} stands as a prominent example in this category. 

This method is based on Mahalanobis distance, which quantifies the distance from a data point to a distribution. The authors assume that pre-trained features extracted from DNN layers can be effectively modelled by a class-conditional Gaussian distribution, given that its posterior distribution is equivalent to the softmax classifier under Gaussian discriminant analysis. Accordingly, for features at a specific layer of the DNN, a Gaussian distribution is fitted for each class using empirical class means and a tied empirical covariance derived from the training samples. Tied covariance denotes a shared covariance matrix across multiple classes in a classification problem, eliminating the need to estimate separate covariance matrices for each class. Instead, a single covariance matrix is computed and shared across all classes. Similarly, class-conditional Gaussian distributions are computed for each layer of the DNN. For instance, in a DNN with $L$ layers trained for a $K$-class classification problem, a total of $L.K$ Gaussian distributions are computed using the training samples.

Once all Gaussian distributions are fitted, OOD testing occurs in several steps. For a selected layer, the Mahalanobis distance from the test input to each class-conditional Gaussian distribution is computed, and the closest class is determined. Subsequently, the test input is perturbed by adding a small, controlled noise calculated based on the Mahalanobis distance to the closest class. Next, the distance from the \textit{perturbed input} to each class-conditional distribution is computed, and the \textit{maximum negative distance} is assigned as the OOD score corresponding to that particular layer. Similarly, OOD scores for all layers are computed, and they are integrated by weighted averaging to derive the final OOD score. The weights are determined by training a logistic regression model using validation samples. Finally, the resulting OOD score is compared with a pre-defined threshold to distinguish OOD/adversarial inputs from clean ID data. While existing methods primarily focus on detecting either OOD or adversarial samples separately, this approach excels in both domains, achieving state-of-the-art performance in the conducted experiments. 

Specifically, in OOD detection, the proposed method outperforms the maximum softmax probability baseline by 5.6\% in terms of AUROC when CIFAR-10 is considered ID, and SVHN, Tiny ImageNet, and LSUN datasets serve as OOD. Furthermore, the authors evaluate the adversarial example detection ability of their method against FGSM~\cite{fgsm}, BIM~\cite{kurakin2018adversarial}, CW~\cite{carlini2017towards}, and DeepFool~\cite{moosavi2016deepfool} attacks. For example, experimental results on a DenseNet~\cite{huang2017densely} trained on CIFAR-10 indicate that the proposed method archives approximately 8\% higher adversarial example detection performance in terms of AUROC compared to the state-of-the-art local intrinsic dimensionality (LID) scores~\cite{ma2018characterizing}.

Similar to the previous method, which employed Mahalanobis distance as the base score, Kaur et al.~\cite{kaur2022idecode} introduce a non-conformity measure (NCM) as the base in their unified robust approach. They propose a method called iDECODe, which stands for In-distribution Equivariance for Conformal Out-of-distribution Detection. iDECODe relies on conformal prediction with transformation equivariance learned from ID data. \textit{Conformal prediction} involves assessing whether a new data point aligns with the training data. In iDECODe, a base model is trained to learn transformation-equivariant representations for a set of transformations $G$ on the training set. Essentially, \textit{equivariance} ensures that the behaviour of the model remains consistent even when the input data is transformed. For instance, when an image is rotated, and a function is applied to it, equivariance guarantees that the output of the function will reflect the same rotation as the input image. Subsequently, for a given input $x$ and a transformation $g \in G$, the NCM is computed as the error between the actual output of the transformed input and the expected output of the transformation-equivariant model for the transformation $g$. Similarly, the authors construct a vector of NCMs using multiple transformations. Finally, these NCMs are aggregated (i.e., summation in this work) to compute a final NCM score, which is then used to determine whether a given test point is OOD or not. The authors also verify that iDECODe guarantees a bounded false positive rate for OOD detection. Furthermore, they empirically show that the same setup can be used to successfully detect adversarial examples generated using FGSM and BIM.

Moreover, Qiu et al.~\cite{qiu2022detecting} introduce the Residual-based Error Detection (RED) framework, which utilises Gaussian processes to estimate the uncertainty of DNN predictions. RED functions as a secondary error detector implemented on top of the base classifier, providing a quantitative measure for identifying OOD and adversarial examples. The fundamental concept behind RED involves employing a Gaussian Process (GP) model to predict the correctness of classification outcomes. Rather than directly aiming to match ground-truth labels, RED focuses on predicting whether the original classification is correct. Accordingly, each training data point is assigned a \textit{target detection score} based on its classification accuracy by the base model. The residual ($e$) between this target score and the original maximum class probability is then computed, and a GP model is trained to predict these residuals. At test time, for a given input $\Bar{x}$, the trained GP model generates a Gaussian distribution for the estimated residual $\Bar{e} \sim \mathcal{N}(\mu_{\Bar{e}},\sigma_{\Bar{e}})$. Here, the mean $\mu_{\Bar{e}}$ serves as an effective separator between correctly and incorrectly classified samples, while the variance $\sigma_{\Bar{e}}$ can be used to for OOD and adversarial example detection. More precisely, RED's detection scores for ID samples exhibit low variance due to their correlation with the training samples. Conversely, samples with high variance indicate uncertainty in RED's detection scores, aiding in the detection of OOD and adversarial samples. 

Similarly, Osada et al.~\cite{osada2023out} propose another reconstruction error-based method for unified robustness, employing \textit{normalising flows}. Normalising flows (NFs) use a series of invertible transformations to map samples from a simple probability distribution (e.g., Gaussian distribution) to a more complex distribution (e.g., the distribution of the data being modelled). The authors further improve the unified robustness by incorporating a \textit{typicality}-based penalty to the NF-based reconstruction error. A typical set is a subset of a larger space containing data that have properties similar to the majority of elements in that space. For instance, if a vector $z$ belongs to the typical set of a $d$-dimensional Gaussian $\mathcal{N}(\mu,\sigma^2 \mathbb{I}_d)$, $z$ satisfies $|z-\mu| \simeq \sigma \sqrt{d}$ with a high probability. This implies that $z$ is concentrated on an annulus centered at $\mu$ with radius $\sigma \sqrt{d}$, which is known as the Gaussian annulus. Leveraging these concepts, the authors propose Penalised Reconstruction Error (PRE) for OOD and adversarial example detection. Firstly, the authors model the data manifold for ID using normalised flows. Inputs lying off this manifold exhibit a high reconstruction error and are flagged as OOD. Secondly, the authors compute the typicality error to measure the distance from an input to the center of the Gaussian annulus. PRE combines these errors to formulate a final score, which is then compared with a pre-defined threshold for OOD and adversarial example detection. 

While previously discussed work employs relatively complex techniques, Lee et al.\cite{lee2022gradient} and Gorbett et al.\cite{gorbett2022utilizing} have opted for simpler approaches to achieve unified robustness. Both of these studies employ a similar approach where they extract information from the primary model and train a secondary classifier to detect OOD and adversarial examples. Subsequently, the prediction score of this secondary classifier is used to determine whether a given input is OOD or adversarial. More precisely, Lee et al.\cite{lee2022gradient} use gradients extracted from the primary classifier as input for their secondary classifier, a multi-layer perceptron (MLP) network. Similarly, Gorbett et al.\cite{gorbett2022utilizing} utilise penultimate and final layer outputs from the primary model as input features for their secondary model, a linear support vector machine (SVM). Furthermore, Meinke et al.~\cite{meinke2019towards} propose a probabilistic model to provide a certifiable guarantee for predictions provided by ReLU models. This method, denoted by \textbf{C}ertified \textbf{C}ertain \textbf{U}ncertainty (CCU), is based on a Gaussian mixture model (GMM) to estimate the centroids and the variances of the distributions. The guarantee provided by this model is twofold. The first guarantee says the classifier has provably low confidence for the OOD data point, far from the training dataset. The second provides an upper bound for the confidence ball around the given input.

\subsection{Other}
\label{subsec:other_unified}

Similar to Section~\ref{subsec:other_robust}, this section explores methods that diverge from the previously discussed categories in Section~\ref{sec:unified_robustness}. These approaches include parameter modelling of DNNs~\cite{ratzlaff2019hypergan} and novel DNN models~\cite{sensoy2020uncertainty}.  

For example, Ratzlaff et al.~\cite{ratzlaff2019hypergan} introduce a generative model named \textit{HyperGAN}, aimed at approximating the posterior distribution of DNN parameters for a target architecture. The methodology involves sampling from a simple multi-dimensional Gaussian distribution and subsequently transforming this sample into multiple vectors using a \textit{mixer}. Each vector is then further transformed into parameters for a corresponding DNN layer by individual generators. An inherent challenge in generating DNN parameters lies in preserving the network's connectivity, where the output of one layer serves as the input to the next. The mixer ensures correlation between the parameters of different layers, which would otherwise be independent. With HyperGAN, the authors can easily sample the generators to produce as many models as required for a particular task.

At test time, the authors assess uncertainty on OOD data by calculating the total predictive entropy from ensembles generated by HyperGAN. They expect that softmax probabilities on ID data will exhibit minimal entropy, with one dominant activation close to 1. Conversely, on OOD data, they expect probabilities to be equally distributed across predictions. To detect adversarial examples, the authors propose that a single adversarial instance will not deceive all parameter variations learned by HyperGAN. By evaluating adversarial examples across multiple generated networks, they observe high entropy among predictions for any individual class.

As another generative approach, Sensoy et al.~\cite{sensoy2020uncertainty} introduce Generative Evidential Neural Networks (GEN), a model capable of expressing both aleatoric and epistemic uncertainty to distinguish decision boundaries and OOD regions in the feature space. They consider the output of the neural network as parameters of a Dirichlet distribution $D(p|\langle\alpha_1,\alpha_2,...,\alpha_K\rangle)$ with a uniform prior instead of a categorical distribution over labels. These parameters correspond to pseudo counts, representing the number of observations or evidence for each class. Furthermore, the authors propose a generative adversarial network (GAN) to generate the most informative OOD samples. These samples are used to learn an implicit density model of the training data, from which pseudo counts (i.e., evidence) are derived for Dirichlet parameters. 

The evidence vector ($e$) for an input $x$ is computed as $e = \exp(f(x;\theta))$, where $f(x;\theta)$ represents the logit vector. Given the uniform prior assumption (i.e., $D(p|\langle1,...,1\rangle)$), the evidence $e_i$ for class $i$ must be added to the respective parameter of this prior (i.e., $\alpha_i = 1+e_i$) to generate the predicted Dirichlet distribution for the sample. If an input $x$ is either OOD or adversarial, GEN is trained to generate minimal evidence (i.e., $e \approx 0$), resulting in a predicted Dirichlet distribution that closely resembles the uniform Dirichlet distribution (indicating high uncertainty). Conversely, for samples labelled as class $k$ in the training set that are not outliers, the authors expect $e_k > e_j \geq 0$ for any $j \neq k$. Empirical results on MNIST and CIFAR-10 datasets demonstrate that GEN significantly improves upon the state-of-the-art in two uncertainty estimation benchmarks: i) detection of OOD samples and ii) robustness to adversarial examples.

\subsection{Summary of Unified Robustness} 

Under unified robustness, we first examined data augmentation-based methods that operate before the model training stage. Typically, these approaches create new training instances by mixing data within the training set to provide a more comprehensive representation of the training distribution. They aim to distinguish distributional shifts, such as adversarial examples and OOD inputs. Subsequently, we analysed learning-based techniques, which intervene during model training. These methods aim to estimate the uncertainty in predictions by employing diverse learning strategies, including self-supervision, novel loss functions, or additional regularisation terms like energy. Following that, we investigated post-hoc score-based methods that extract information from pre-trained models to differentiate between clean ID, adversarial ID, and clean OOD instances. These methods formulate detection scores based on metrics such as distance, non-conformity, residual errors, or prediction confidence from secondary classifiers. Lastly, we explored approaches that are capable of detecting both OOD and adversarial inputs, which do not fit directly into any of the previously mentioned categories. These methods included modelling the parameter space of target DNN architectures through generative networks and proposing innovative DNN architectures like generative evidential networks. Table~\ref{tab:uni_rob_summary} presents a summary of the methods on unified robustness.

\begin{center}
\begin{boxM}
\footnotesize
\centering \textbf{\underline{Unified robustness}} 
\vspace{10pt}
\begin{itemize}

\item \textbf{Data augmentation-based} methods create new training instances by mixing training data with complex structures to better represent the training distribution. 
    \begin{itemize}
        \item \textbf{Pros}: Better generalisability, Examine the ID distribution comprehensively through augmented data.
        \item \textbf{Cons}: Augmented data may not fully capture the complexity and variability in real data, Risk of over-fitting.
    \end{itemize}
\item \textbf{Learning-based} techniques estimate uncertainty in predictions, utilising diverse training strategies.
    \begin{itemize}
        \item \textbf{Pros}: Specifically trained to improve uncertainty estimates.
        \item \textbf{Cons}: Increased training time due to advanced training approaches.
    \end{itemize}
\item \textbf{Score-based} post-hoc methods extract information from pre-trained models to formulate detection scores to distinguish adversarial ID and OOD inputs.
    \begin{itemize}
        \item \textbf{Pros}: Easy implementation, High performance on clean ID data.
        \item \textbf{Cons}: Inherits any biases present in the pre-trained model, May require additional fine-tuning.
    \end{itemize}
\item \textbf{Other} approaches detect both OOD and adversarial ID inputs through modelling the parameter space of target DNNs and novel DNN architectures.
    \begin{itemize}
        \item \textbf{Pros}: Specialised for uncertainty estimation, Easy model ensemble generation. 
        \item \textbf{Cons}: Limited scalability.
    \end{itemize}
\end{itemize}
\end{boxM}
\end{center}

\begin{table*}[t!]
\scriptsize
  \centering
  \caption{\centering Summary of unified robust methods. \break Note that OOD datasets and performance are reported for the ID dataset marked in \textbf{bold}.} \vspace{-2mm}
  \label{tab:uni_rob_summary}
  \resizebox{\textwidth}{!}{%
    \begin{tabular}{p{3cm}|p{5.0cm}|p{2cm}|p{2.2cm}|p{1.5cm}|p{2.5cm}|p{2.5cm}}
      \toprule
      & & & & & \multicolumn{2}{c}{\textbf{Performance}} \\
      \textbf{Research Work} & \textbf{Summary} & \textbf{ID datasets} & \textbf{OOD datasets} & \textbf{Adversary} & \textbf{OOD detection} & \textbf{Adversarial\break robustness} \\
      \midrule
        \multicolumn{6}{c}{\textbf{Data augmentation-based}} \\
    \midrule
      \midrule
      Pinto et al.~\cite{pinto2022using} & Generate new data points by combining two training instances linearly. A regulariser on the new samples is added to the training criteria.  & \textbf{CIFAR-10}, CIFAR-100, ImageNet & SVHN, CIFAR-100, TinyImageNet & Corrupted CIFAR-10 & AUROC - 92.18\% & Detection rate - 83.13\% \\
      \midrule
      Hendrycks et al.~\cite{hendrycks2022pixmix} & Augments the dataset by creating new instances by mixing training data with complex structures. & \textbf{CIFAR-10}, CIFAR-100, ImageNet & SVHN, Textures, LSUN, Places69& $l_{\infty}$-PGD \break $\epsilon = 2/255$ & AUROC - 97.00\% & Detection error - 82.10\% \\
      \midrule
      Azizmalayeri et al.~\cite{azizmalayeri2023data} & Detect and 
      remove hard samples from the training process rather than applying complicated algorithms to mitigate their effects. & \textbf{CIFAR-10}, SVHN & CIFAR-100 & $l_{\infty}$-PGD \break $\epsilon = 8/255$ & Detection rate - 80.00\% & Robust accuracy - 49.74\% \\
    \midrule
    \multicolumn{6}{c}{\textbf{Learning-based}} \\
    \midrule
    \midrule
      Hendrycks et al.~\cite{hendrycks2019using_self_supervised} & Demonstrate that self-supervision can improve robustness to adversarial examples and OOD data. & \textbf{CIFAR-10} & SVHN, Textures, Places365, LSUN, CIFAR-100 & $l_{\infty}$-PGD \break $\epsilon = 8/255$ & AUROC - 96.20\% & Robust accuracy - 50.40\% \\
      \midrule
      Hendrycks et al.~\cite{hendrycks2019using_pre-trained} & Demonstrate that adversarial pre-training can improve model robustness and uncertainty estimates. & \textbf{CIFAR-10}, CIFAR-100 & Textures, Places365 & $l_{\infty}$-PGD \break $\epsilon = 8/255$ & AUROC - 94.50\% & Robust accuracy - 57.40\%\\
      \midrule
      Malinin et al.~\cite{malinin2019reverse} & Propose reverse KL divergence as the training criterion for prior networks to overcome scalability limitations and improve uncertainty estimates. & \textbf{CIFAR-10}, CIFAR-100 & SVHN, LSUN, TinyImageNet & $l_{\infty}$-MIM \break $\epsilon = 30/128$ & AUROC - 96.53\% & ASR - 25.00\%\\
      \midrule
      Lee et al.~\cite{lee2020adversarial} & Propose a hybrid model based on adversarial training and joint energy-based modelling for effective OOD and adversarial example detection. & \textbf{CIFAR-10} & SVHN & $l_{\infty}$-PGD \break $\epsilon = 8/255$ & AUROC - 87.00\% & Robust accuracy - 52.36\%\\
    \midrule
    \multicolumn{6}{c}{\textbf{Score-based}} \\
    \midrule
    \midrule
      Lee et al.~\cite{lee2018simple} & Propose a score based on Mahalanobis distance to detect anomalous inputs such as adversarial and OOD inputs. & \textbf{CIFAR-10}, CIFAR-100, SVHN & SVHN, LSUN, TinyImageNet & $l_{\infty}$-FGSM, BIM, CW, DeepFool & TNR95 - 94.33\% \break AUROC - 98.73\% & AUROC - 92.71\% \\
      \midrule
      Kaur et al.~\cite{kaur2022idecode} & Propose a base non-conformity measure and a new aggregation method leveraging ID equivariance for conformal OOD and adversarial example detection. & \textbf{CIFAR-10} & SVHN, LSUN, ImageNet, CIFAR100, Places365 & $l_{\infty}$-FGSM, BIM, CW, DeepFool & AUROC - 89.53\% & AUROC - 95.63\% \\
      \midrule
      Qiu et al.~\cite{qiu2022detecting} & Propose a framework that quantifies misclassification errors using an error detector which estimates detection score uncertainty via Gaussian Processes. & \textbf{CIFAR-10} & SVHN & FGSM & AUPR - 86.28\% & AUPR $\approx$ 75.00\% \\
      \midrule
      Osada et al.~\cite{osada2023out} & Propose a novel reconstruction error based on normalising flows, further enhanced using a typicality-based penalty. & \textbf{CIFAR-10}, TinyImageNet, ImageNet & CelebA, LSUN, TinyImageNet & $l_{\infty}$-PGD \break $\epsilon = 8/255$ \break \break $l_{2}$-CW & AUROC - 96.04\% & \textbf{PGD}\break AUROC - 99.93\% \break \textbf{CW} \break AUROC - 95.11\% \\
      \midrule
       Lee et al.~\cite{lee2022gradient} & Train a secondary classifier to detect adversarial and OOD inputs using gradient information at different layers of the primary model. & \textbf{CIFAR-10}, SVHN & SVHN, LSUN, TinyImageNet & FGSM, CW, PGD, BIM & AUROC - 98.78\% & AUROC - 95.63\% \\
       \midrule
      Gorbett et al.~\cite{gorbett2022utilizing} &  Train a linear SVM to detect four types of erroneous inputs using the hidden and softmax feature vectors of pre-trained neural networks. & \textbf{CIFAR-10}, TinyImageNet, ImageNet & SVHN, SUN, CIFAR-100 & FGSM, CW, PGD & AUROC - 98.63\% & AUROC - 99.28\% \\
      \midrule
      Meinke et al.~\cite{meinke2019towards} & Propose a Gaussian mixture model-based probabilistic model to provide a certifiable guarantee for predictions provided by ReLU models. & \textbf{CIFAR-10}, CIFAR-100, MNIST, SVHN, FMNIST  & SVHN, CIFAR-100, LSUN(crop), ImageNet & $l_{\infty}$-PGD & AUROC - 96.82\% & AUROC - 100.00\%  \\
      \midrule
          \multicolumn{6}{c}{\textbf{Other}} \\
    \midrule
    \midrule
       Ratzlaff et al.~\cite{ratzlaff2019hypergan} & A generative model, HyperGAN to learn a distribution of model parameters. It provides better uncertainty estimates on OOD and adversarial inputs. & \textbf{MNIST}, CIFAR-10 (first 5 classes) & notMNIST & FGSM, PGD & Numerical results are not reported. & Numerical results are not reported.\\
       \midrule
       Sensoy et al.~\cite{sensoy2020uncertainty} & A generative model capturing both aleatoric and epistemic uncertainty, distinguishing decision boundaries and OOD regions in the feature space. & \textbf{CIFAR-10} \break (first 5 classes), MNIST & CIFAR-10 \break (last 5 classes) & FGSM & AUROC - 77.50\% & Numerical results are not reported. \\
      \bottomrule
    \end{tabular}%
  }
\end{table*} \vspace{-1mm}

\section{Limitations and Future Directions}
\label{sec:limitations_future}

Next, we highlight the limitations and challenges associated with existing methods for robust OOD detection and unified robustness and discuss potential future research directions.

\subsection{Lack of benchmarks}

One significant issue observed in robust OOD detection, as indicated by Table~\ref{tab:rob_ood_summary}, is the absence of a standard evaluation strategy akin to the OpenOOD benchmarking library~\cite{zhang2023openood} used in standard OOD detection or AutoAttack library~\cite{croce2020reliable} for adversarial robustness. Instead, various robust OOD work assess the robustness of OOD detectors using different attack criteria. This lack of consistency makes it challenging to determine the current State-of-the-Art (SOTA) and accurately compare different methods to understand the robustness of the current detectors. Consequently, there is a pressing need for a standardised approach to evaluate the robustness of OOD detectors.

\subsection{Certified defences}

The main focus in existing research has been on building empirically robust OOD detectors, with a limited exploration into provable guarantees for robust OOD detection~\cite{bitterwolf2020certifiably,meinke2022provably}. However, empirical detectors are more vulnerable to sophisticated adaptive attacks compared to provable detectors. This is analogous to regular \textit{adversarial robustness}, where provable defences are preferred over empirical defences~\cite{carlini2019evaluating}. For example, defensive distillation~\cite{papernot2016distillation}, a well-known empirical defence against adversarial examples, is compromised by more advanced Carlini-Wagner attack~\cite{carlini2017towards}. Therefore, in light of tightened AI regulations, it is imperative for future research efforts to prioritise the development of provably robust OOD detectors, particularly aiming for real-world safety-critical applications. 

Nonetheless, one known drawback of provable defences in the context of regular adversarial robustness is that they only provide guarantees for a specific set of inputs~\cite{carlini2019evaluating}. For example, such guarantees take the form \textit{"for some specific set of inputs $x \in \mathcal{X}$, no adversarial examples with distortion less than $\epsilon$ exist."}. Usually, these methods offer no assurances about any other inputs $x' \notin \mathcal{X}$. It is more likely that the same limitations exist, even more so in the context of robust OOD detection.

\subsection{Scale and depth of attacks}

As can be seen from Table~\ref{tab:uni_rob_summary}, many existing studies on unified robustness have conducted experiments only on small-scale datasets such as MNIST and CIFAR-10~\cite{lee2020adversarial,ratzlaff2019hypergan,sensoy2020uncertainty,qiu2022detecting}. However, these methods are not often evaluated on complex datasets such as Tiny ImageNet~\cite{Le2015TinyIV} and ImageNet-1K~\cite{deng2009imagenet}, possibly due to scalability limitations such as computational complexity and memory requirements. As a result, their applicability in real-world scenarios remains uncertain. Therefore, future research should prioritise scalability when proposing robust approaches and evaluate them on complex datasets containing a large number of classes.

Furthermore, in many existing studies, robustness is often achieved against specific OOD or adversarial inputs. Particularly, in outlier exposure-based robust OOD detection methods, the process of sampling outliers for training is prohibitively large, leading to trained models exhibiting bias towards the seen outliers. In contrast, a few existing works~\cite{dionelis2022frob,yin2022learning} attempt to model the support of ID data using the ID training set and a limited number of outliers near the support boundary. This approach helps avoid generalisation limitations caused by bias towards seen outliers during training. Similar to such attempts, future work must place more emphasis on developing methods that effectively model the support of ID data rather than exposing models only to a subset of OOD inputs. By effectively modelling the support of ID data, robust OOD detectors can be constructed, leveraging knowledge of the complement of the support.

Similarly, most evaluations against adversarial inputs are confined to PGD adversaries with lower perturbation strengths. As a result, the robustness achieved by the discussed methods may be limited. Additionally, existing research mainly focuses on introducing constrained perturbations for OOD inputs to assess the adversarial robustness of OOD detectors. However, adversarial OOD inputs do not necessarily need to be constrained, as they can vary widely in appearance. Analogously, in the context of conventional adversarial attacks, Carlini et al.~\cite{carlini2017adversarial} showed that adversarial examples can be created using larger perturbations while still remaining imperceptible. Consequently, attackers can freely modify OOD inputs without being restricted by norm constraints. Defending against such adversarial examples poses challenges for techniques like adversarial training, which typically defends against attacks within an $\epsilon$-ball around an OOD input. 

\subsection{Unified handling of OOD and adversarial examples}

Investigating OOD and adversarial inputs within a unified framework presents an exciting research direction where it is possible to build a more robust input pre-filtering pipeline that can be used to collectively mitigate the threat of both adversarial and OOD inputs.

\begin{figure}[t!]
\centering
\includegraphics[width=0.56\textwidth]{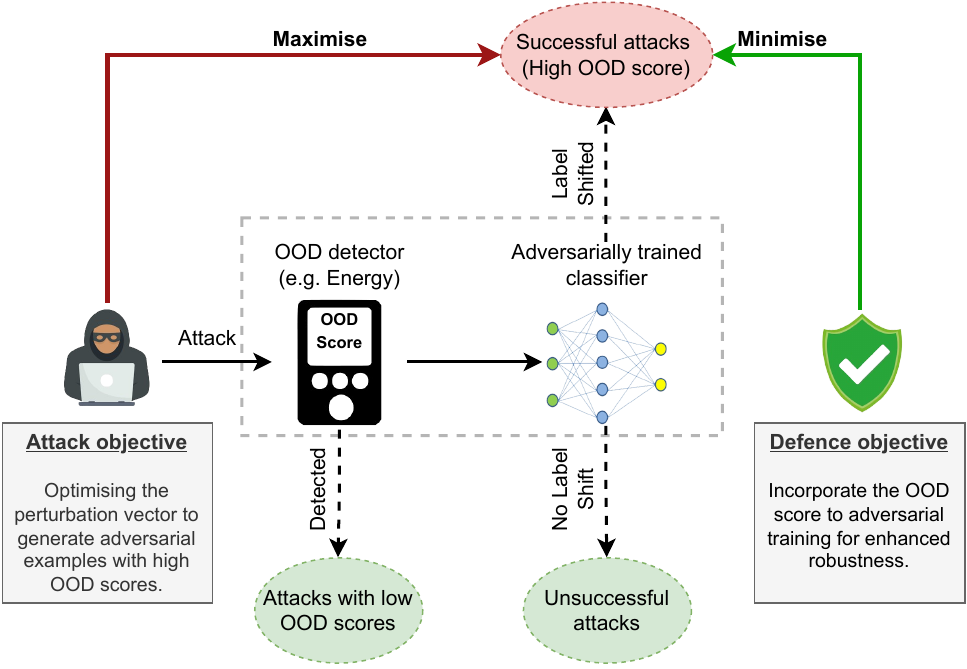} 
\caption{Unified defence system} \vspace{-6mm}
\label{fig:unified_ml_system}
\end{figure} 

For example, Figure~\ref{fig:unified_ml_system} illustrates a unified defence system comprising an \textit{OOD detector} and an \textit{adversarially trained classifier}. Here, the attacker's objective is to deceive the primary classifier. In this configuration, adversarial examples generated by the attacker may fail for two main reasons. Firstly, if the adversarial input exhibits a low OOD score ({\bf cf.} Equation~\ref{eq:ood_detection}), it will be discarded by the OOD detector. Secondly, even if some adversarial examples manage to bypass the OOD detector, they may still fail to alter the predicted label due to the robustness achieved through adversarial training, rendering them unsuccessful. Therefore, the attack's success rate can be lowered when an OOD detector is used in conjunction with adversarial training. Furthermore, there is an opportunity for adversarial defences and OOD detectors to collaborate to enhance the overall robustness of DNNs. Since both adversarial perturbations and OOD inputs represent different types of distributional shifts, the OOD detector can flag suspicious, uncertain inputs for the adversarial detector or the primary classifier while discarding definite OOD inputs. Through such knowledge-sharing and mutual reinforcement strategies, it becomes feasible to enhance the overall robustness of DNNs against these distribution shifts. 
 
On the other hand, against such a unified defence, attackers must optimise the perturbation vector not only to deceive the primary classifier but also to evade OOD detection. Adversarial attacks typically compute the minimum perturbation required to execute the attack. For instance, in an untargeted attack, this minimum perturbation alters the predicted label, while in a targeted attack, it classifies the input into the target class. Accordingly, the maximum softmax probability or maximum logit of the adversarial example is often lower than that of the original input. As a result, while adversarial examples may deceive the primary classifier, they are highly likely to be detected even by a simple OOD detector, such as those based on maximum softmax probability (MSP)~\cite{hendrycks2016baseline} or energy~\cite{energy}. 
However, against simple OOD detectors, a potential strategy to evade detection is to continue perturbing the input (i.e., exceeding the minimum required perturbation) until the adversarial example is highly confident about an incorrect or the target class. This approach prevents from being flagged by an energy or MSP OOD detector. However, if a more sophisticated OOD detector, such as DICE~\cite{dice} or ReAct~\cite{react}, is employed, evading detection may not be as straightforward. Hence, an opportunity exists to optimise adversarial perturbations to circumvent a unified defence comprising both OOD detection and adversarial robustness.

There is an emerging research area known as \textit{full-spectrum OOD detection} (FS-OOD), which closely aligns with the concept of unified defence. FS-OOD~\cite{yang2023full,lu2023likelihood} aims to detect both covariate and semantic shifts using a single method. However, these methods are not evaluated on the worst-case covariate shifts, i.e., adversarial examples. Thus, unified defence can be considered a broader and challenging problem which has the ability to mitigate all forms of distributional shifts.

\section{Conclusion}
\label{conclusion}

In this paper, we survey research at the intersection of adversarial examples and OOD inputs, exploring how they are studied together in literature. Accordingly, we identified two key research areas: robust OOD detection and unified robustness. Furthermore, we introduced a novel taxonomy centered around distributional shifts, positioning these areas alongside other related research problems. We systematically reviewed existing work on robust OOD detection and unified robustness, discussing methodologies and their strengths and weaknesses across four sub-categories. Lastly, we highlighted the limitations of current research and proposed potential improvements and future research directions that explore adversarial and OOD inputs within a unified framework to ensure the reliable operation of DNNs in real-world applications.

\bibliographystyle{ACM-Reference-Format}
\bibliography{bibliography}


\begin{thebibliography}{121}

%%% the Web address.

\ifx \showCODEN    \undefined \def \showCODEN     #1{\unskip}     \fi
\ifx \showDOI      \undefined \def \showDOI       #1{#1}\fi
\ifx \showISBNx    \undefined \def \showISBNx     #1{\unskip}     \fi
\ifx \showISBNxiii \undefined \def \showISBNxiii  #1{\unskip}     \fi
\ifx \showISSN     \undefined \def \showISSN      #1{\unskip}     \fi
\ifx \showLCCN     \undefined \def \showLCCN      #1{\unskip}     \fi
\ifx \shownote     \undefined \def \shownote      #1{#1}          \fi
\ifx \showarticletitle \undefined \def \showarticletitle #1{#1}   \fi
\ifx \showURL      \undefined \def \showURL       {\relax}        \fi
% invisible to TeX
\providecommand\bibfield[2]{#2}
\providecommand\bibinfo[2]{#2}
\providecommand\natexlab[1]{#1}
\providecommand\showeprint[2][]{arXiv:#2}

\bibitem[whi({[n.\,d.]})]%
        {whitehouseBlueprintBill}
 \bibinfo{year}{[n.\,d.]}\natexlab{}.
\newblock \bibinfo{title}{{B}lueprint for an {A}{I} {B}ill of {R}ights | {O}{S}{T}{P} | {T}he {W}hite {H}ouse --- whitehouse.gov}.
\newblock \bibinfo{howpublished}{\url{https://www.whitehouse.gov/ostp/ai-bill-of-rights/}}.
\newblock


\bibitem[art({[n.\,d.]})]%
        {artificialintelligenceactHome}
 \bibinfo{year}{[n.\,d.]}\natexlab{}.
\newblock \bibinfo{title}{{H}ome --- artificialintelligenceact.eu}.
\newblock \bibinfo{howpublished}{\url{https://artificialintelligenceact.eu/}}.
\newblock


\bibitem[Aldahdooh et~al\mbox{.}(2022)]%
        {adversarial_survey}
\bibfield{author}{\bibinfo{person}{Ahmed Aldahdooh}, \bibinfo{person}{Wassim Hamidouche}, \bibinfo{person}{Sid~Ahmed Fezza}, {and} \bibinfo{person}{Olivier D{\'{e}}forges}.} \bibinfo{year}{2022}\natexlab{}.
\newblock \showarticletitle{Adversarial example detection for {DNN} models: a review and experimental comparison}.
\newblock \bibinfo{journal}{\emph{Artificial Intelligence Review}} \bibinfo{volume}{55}, \bibinfo{number}{6} (\bibinfo{year}{2022}).
\newblock


\bibitem[Aryal et~al\mbox{.}({[n.\,d.]})]%
        {aryal2111survey}
\bibfield{author}{\bibinfo{person}{K Aryal}, \bibinfo{person}{M Gupta}, {and} \bibinfo{person}{M Abdelsalam}.} \bibinfo{year}{[n.\,d.]}\natexlab{}.
\newblock \showarticletitle{A survey on adversarial attacks for malware analysis. arXiv 2021}.
\newblock \bibinfo{journal}{\emph{arXiv preprint arXiv:2111.08223}} (\bibinfo{year}{[n.\,d.]}).
\newblock


\bibitem[Augustin et~al\mbox{.}(2020)]%
        {augustin2020adversarial}
\bibfield{author}{\bibinfo{person}{Maximilian Augustin}, \bibinfo{person}{Alexander Meinke}, {and} \bibinfo{person}{Matthias Hein}.} \bibinfo{year}{2020}\natexlab{}.
\newblock \showarticletitle{Adversarial robustness on in-and out-distribution improves explainability}. In \bibinfo{booktitle}{\emph{European Conference on Computer Vision}}. Springer, \bibinfo{pages}{228--245}.
\newblock


\bibitem[Azizmalayeri et~al\mbox{.}(2022)]%
        {azizmalayeri2022your}
\bibfield{author}{\bibinfo{person}{Mohammad Azizmalayeri}, \bibinfo{person}{Arshia Soltani~Moakhar}, \bibinfo{person}{Arman Zarei}, \bibinfo{person}{Reihaneh Zohrabi}, {et~al\mbox{.}}} \bibinfo{year}{2022}\natexlab{}.
\newblock \showarticletitle{Your Out-of-Distribution Detection Method is Not Robust!}
\newblock \bibinfo{journal}{\emph{Advances in Neural Information Processing Systems}}  \bibinfo{volume}{35} (\bibinfo{year}{2022}), \bibinfo{pages}{4887--4901}.
\newblock


\bibitem[Azizmalayeri et~al\mbox{.}(2023)]%
        {azizmalayeri2023data}
\bibfield{author}{\bibinfo{person}{Mohammad Azizmalayeri}, \bibinfo{person}{Arman Zarei}, \bibinfo{person}{Alireza Isavand}, {et~al\mbox{.}}} \bibinfo{year}{2023}\natexlab{}.
\newblock \showarticletitle{A Data-Centric Approach for Improving Adversarial Training Through the Lens of Out-of-Distribution Detection}. In \bibinfo{booktitle}{\emph{2023 28th CSICC}}. IEEE, \bibinfo{pages}{1--6}.
\newblock


\bibitem[Ben-David et~al\mbox{.}(2010)]%
        {covariate_1}
\bibfield{author}{\bibinfo{person}{Shai Ben-David}, \bibinfo{person}{John Blitzer}, \bibinfo{person}{Koby Crammer}, \bibinfo{person}{Alex Kulesza}, \bibinfo{person}{Fernando Pereira}, {and} \bibinfo{person}{Jennifer Vaughan}.} \bibinfo{year}{2010}\natexlab{}.
\newblock \showarticletitle{A theory of learning from different domains}.
\newblock \bibinfo{journal}{\emph{Machine Learning}}  \bibinfo{volume}{79} (\bibinfo{year}{2010}), \bibinfo{pages}{151--175}.
\newblock


\bibitem[Berrada et~al\mbox{.}(2021)]%
        {berrada2021make}
\bibfield{author}{\bibinfo{person}{Leonard Berrada}, \bibinfo{person}{Sumanth Dathathri}, \bibinfo{person}{Krishnamurthy Dvijotham}, {et~al\mbox{.}}} \bibinfo{year}{2021}\natexlab{}.
\newblock \showarticletitle{Make sure you're unsure: A framework for verifying probabilistic specifications}.
\newblock \bibinfo{journal}{\emph{Advances in Neural Information Processing Systems 2021}} (\bibinfo{year}{2021}).
\newblock


\bibitem[Bilge and Dumitra\c{s}(2012)]%
        {zero_day_attacks}
\bibfield{author}{\bibinfo{person}{Leyla Bilge} {and} \bibinfo{person}{Tudor Dumitra\c{s}}.} \bibinfo{year}{2012}\natexlab{}.
\newblock \showarticletitle{Before We Knew It: An Empirical Study of Zero-Day Attacks in the Real World}. In \bibinfo{booktitle}{\emph{Proceedings of the 2012 ACM Conference on Computer and Communications Security}} \emph{(\bibinfo{series}{CCS '12})}. \bibinfo{pages}{833–844}.
\newblock
\showISBNx{9781450316514}


\bibitem[Bitterwolf et~al\mbox{.}(2020)]%
        {bitterwolf2020certifiably}
\bibfield{author}{\bibinfo{person}{Julian Bitterwolf}, \bibinfo{person}{Alexander Meinke}, {and} \bibinfo{person}{Matthias Hein}.} \bibinfo{year}{2020}\natexlab{}.
\newblock \showarticletitle{Certifiably adversarially robust detection of out-of-distribution data}.
\newblock \bibinfo{journal}{\emph{Advances in Neural Information Processing Systems}}  \bibinfo{volume}{33} (\bibinfo{year}{2020}), \bibinfo{pages}{16085--16095}.
\newblock


\bibitem[Bulatov(2011)]%
        {bulatov2011notmnist}
\bibfield{author}{\bibinfo{person}{Yaroslav Bulatov}.} \bibinfo{year}{2011}\natexlab{}.
\newblock \showarticletitle{Notmnist dataset}.
\newblock \bibinfo{journal}{\emph{Google (Books/OCR), Tech. Rep.[Online]. Available: http://yaroslavvb. blogspot. it/2011/09/notmnist-dataset. html}}  \bibinfo{volume}{2} (\bibinfo{year}{2011}).
\newblock


\bibitem[Carlini et~al\mbox{.}(2019)]%
        {carlini2019evaluating}
\bibfield{author}{\bibinfo{person}{Nicholas Carlini}, \bibinfo{person}{Anish Athalye}, \bibinfo{person}{Nicolas Papernot}, \bibinfo{person}{Wieland Brendel}, \bibinfo{person}{Jonas Rauber}, \bibinfo{person}{Dimitris Tsipras}, \bibinfo{person}{Ian Goodfellow}, \bibinfo{person}{Aleksander Madry}, {and} \bibinfo{person}{Alexey Kurakin}.} \bibinfo{year}{2019}\natexlab{}.
\newblock \showarticletitle{On evaluating adversarial robustness}.
\newblock \bibinfo{journal}{\emph{preprint arXiv:1902.06705}} (\bibinfo{year}{2019}).
\newblock


\bibitem[Carlini and Wagner(2017a)]%
        {carlini2017adversarial}
\bibfield{author}{\bibinfo{person}{Nicholas Carlini} {and} \bibinfo{person}{David Wagner}.} \bibinfo{year}{2017}\natexlab{a}.
\newblock \showarticletitle{Adversarial examples are not easily detected: Bypassing ten detection methods}. In \bibinfo{booktitle}{\emph{Proceedings of the 10th ACM workshop on artificial intelligence and security}}. \bibinfo{pages}{3--14}.
\newblock


\bibitem[Carlini and Wagner(2017b)]%
        {carlini2017towards}
\bibfield{author}{\bibinfo{person}{Nicholas Carlini} {and} \bibinfo{person}{David Wagner}.} \bibinfo{year}{2017}\natexlab{b}.
\newblock \showarticletitle{Towards evaluating the robustness of neural networks}. In \bibinfo{booktitle}{\emph{2017 ieee symposium on security and privacy (sp)}}. Ieee, \bibinfo{pages}{39--57}.
\newblock


\bibitem[Chandola et~al\mbox{.}(2009)]%
        {chandola2009anomaly}
\bibfield{author}{\bibinfo{person}{Varun Chandola}, \bibinfo{person}{Arindam Banerjee}, {and} \bibinfo{person}{Vipin Kumar}.} \bibinfo{year}{2009}\natexlab{}.
\newblock \showarticletitle{Anomaly detection: A survey}.
\newblock \bibinfo{journal}{\emph{ACM computing surveys (CSUR)}} \bibinfo{volume}{41}, \bibinfo{number}{3} (\bibinfo{year}{2009}), \bibinfo{pages}{1--58}.
\newblock


\bibitem[Chapelle et~al\mbox{.}(2000)]%
        {chapelle2000vicinal}
\bibfield{author}{\bibinfo{person}{Olivier Chapelle}, \bibinfo{person}{Jason Weston}, \bibinfo{person}{L{\'e}on Bottou}, {and} \bibinfo{person}{Vladimir Vapnik}.} \bibinfo{year}{2000}\natexlab{}.
\newblock \showarticletitle{Vicinal risk minimization}.
\newblock \bibinfo{journal}{\emph{Advances in neural information processing systems}}  \bibinfo{volume}{13} (\bibinfo{year}{2000}).
\newblock


\bibitem[Chen et~al\mbox{.}(2020b)]%
        {chen2020robust}
\bibfield{author}{\bibinfo{person}{Jiefeng Chen}, \bibinfo{person}{Yixuan Li}, \bibinfo{person}{Xi Wu}, \bibinfo{person}{Yingyu Liang}, {and} \bibinfo{person}{Somesh Jha}.} \bibinfo{year}{2020}\natexlab{b}.
\newblock \showarticletitle{Robust out-of-distribution detection for neural networks}.
\newblock \bibinfo{journal}{\emph{arXiv preprint arXiv:2003.09711}} (\bibinfo{year}{2020}).
\newblock


\bibitem[Chen et~al\mbox{.}(2021)]%
        {chen2021atom}
\bibfield{author}{\bibinfo{person}{Jiefeng Chen}, \bibinfo{person}{Yixuan Li}, \bibinfo{person}{Xi Wu}, \bibinfo{person}{Yingyu Liang}, {and} \bibinfo{person}{Somesh Jha}.} \bibinfo{year}{2021}\natexlab{}.
\newblock \showarticletitle{Atom: Robustifying out-of-distribution detection using outlier mining}. In \bibinfo{booktitle}{\emph{European Conference, ECML PKDD 2021}}. Springer, \bibinfo{pages}{430--445}.
\newblock


\bibitem[Chen et~al\mbox{.}(2020a)]%
        {chen2020self}
\bibfield{author}{\bibinfo{person}{Kejiang Chen}, \bibinfo{person}{Yuefeng Chen}, \bibinfo{person}{Hang Zhou}, \bibinfo{person}{Xiaofeng Mao}, \bibinfo{person}{Yuhong Li}, \bibinfo{person}{Yuan He}, \bibinfo{person}{Hui Xue}, {et~al\mbox{.}}} \bibinfo{year}{2020}\natexlab{a}.
\newblock \showarticletitle{Self-supervised adversarial training}. In \bibinfo{booktitle}{\emph{IEEE International Conference on Acoustics, Speech and Signal Processing}}.
\newblock


\bibitem[Chen et~al\mbox{.}(2017)]%
        {chen2017zoo}
\bibfield{author}{\bibinfo{person}{Pin-Yu Chen}, \bibinfo{person}{Huan Zhang}, \bibinfo{person}{Yash Sharma}, \bibinfo{person}{Jinfeng Yi}, {and} \bibinfo{person}{Cho-Jui Hsieh}.} \bibinfo{year}{2017}\natexlab{}.
\newblock \showarticletitle{Zoo: Zeroth order optimization based black-box attacks to deep neural networks without training substitute models}. In \bibinfo{booktitle}{\emph{Proceedings of the 10th ACM workshop on artificial intelligence and security}}. \bibinfo{pages}{15--26}.
\newblock


\bibitem[Cimpoi et~al\mbox{.}(2014)]%
        {cimpoi2014describing}
\bibfield{author}{\bibinfo{person}{Mircea Cimpoi}, \bibinfo{person}{Subhransu Maji}, \bibinfo{person}{Iasonas Kokkinos}, \bibinfo{person}{Sammy Mohamed}, {and} \bibinfo{person}{Andrea Vedaldi}.} \bibinfo{year}{2014}\natexlab{}.
\newblock \showarticletitle{Describing textures in the wild}. In \bibinfo{booktitle}{\emph{Proceedings of the IEEE conference on computer vision and pattern recognition}}. \bibinfo{pages}{3606--3613}.
\newblock


\bibitem[Croce and Hein(2020)]%
        {croce2020reliable}
\bibfield{author}{\bibinfo{person}{Francesco Croce} {and} \bibinfo{person}{Matthias Hein}.} \bibinfo{year}{2020}\natexlab{}.
\newblock \showarticletitle{Reliable evaluation of adversarial robustness with an ensemble of diverse parameter-free attacks}. In \bibinfo{booktitle}{\emph{International conference on machine learning}}. PMLR, \bibinfo{pages}{2206--2216}.
\newblock


\bibitem[Cui and Wang(2022)]%
        {cui2022out}
\bibfield{author}{\bibinfo{person}{Peng Cui} {and} \bibinfo{person}{Jinjia Wang}.} \bibinfo{year}{2022}\natexlab{}.
\newblock \showarticletitle{Out-of-distribution (OOD) detection based on deep learning: A review}.
\newblock \bibinfo{journal}{\emph{Electronics}} \bibinfo{volume}{11}, \bibinfo{number}{21} (\bibinfo{year}{2022}), \bibinfo{pages}{3500}.
\newblock


\bibitem[Dahanayaka et~al\mbox{.}(2023)]%
        {dahanayaka2023robust}
\bibfield{author}{\bibinfo{person}{Thilini Dahanayaka}, \bibinfo{person}{Yasod Ginige}, \bibinfo{person}{Yi Huang}, \bibinfo{person}{Guillaume Jourjon}, {and} \bibinfo{person}{Suranga Seneviratne}.} \bibinfo{year}{2023}\natexlab{}.
\newblock \showarticletitle{Robust open-set classification for encrypted traffic fingerprinting}.
\newblock \bibinfo{journal}{\emph{Computer Networks}}  \bibinfo{volume}{236} (\bibinfo{year}{2023}), \bibinfo{pages}{109991}.
\newblock


\bibitem[Davis and Goadrich(2006)]%
        {davis2006relationship}
\bibfield{author}{\bibinfo{person}{Jesse Davis} {and} \bibinfo{person}{Mark Goadrich}.} \bibinfo{year}{2006}\natexlab{}.
\newblock \showarticletitle{The relationship between Precision-Recall and ROC curves}. In \bibinfo{booktitle}{\emph{Proceedings of the 23rd international conference on Machine learning}}. \bibinfo{pages}{233--240}.
\newblock


\bibitem[Deng et~al\mbox{.}(2009)]%
        {deng2009imagenet}
\bibfield{author}{\bibinfo{person}{Jia Deng}, \bibinfo{person}{Wei Dong}, \bibinfo{person}{Richard Socher}, \bibinfo{person}{Li-Jia Li}, \bibinfo{person}{Kai Li}, {and} \bibinfo{person}{Li Fei-Fei}.} \bibinfo{year}{2009}\natexlab{}.
\newblock \showarticletitle{Imagenet: A large-scale hierarchical image database}. In \bibinfo{booktitle}{\emph{2009 IEEE conference on computer vision and pattern recognition}}. Ieee, \bibinfo{pages}{248--255}.
\newblock


\bibitem[Deng(2012)]%
        {deng2012mnist}
\bibfield{author}{\bibinfo{person}{Li Deng}.} \bibinfo{year}{2012}\natexlab{}.
\newblock \showarticletitle{The mnist database of handwritten digit images for machine learning research}.
\newblock \bibinfo{journal}{\emph{IEEE Signal Processing Magazine}} \bibinfo{volume}{29}, \bibinfo{number}{6} (\bibinfo{year}{2012}), \bibinfo{pages}{141--142}.
\newblock


\bibitem[Dionelis et~al\mbox{.}(2022)]%
        {dionelis2022frob}
\bibfield{author}{\bibinfo{person}{Nikolaos Dionelis}, \bibinfo{person}{Sotirios~A Tsaftaris}, {and} \bibinfo{person}{Mehrdad Yaghoobi}.} \bibinfo{year}{2022}\natexlab{}.
\newblock \showarticletitle{FROB: Few-Shot ROBust Model for Joint Classification and Out-of-Distribution Detection}. In \bibinfo{booktitle}{\emph{Joint European Conference on Machine Learning and Knowledge Discovery in Databases}}. Springer, \bibinfo{pages}{137--153}.
\newblock


\bibitem[Dong et~al\mbox{.}(2018)]%
        {dong2018boosting}
\bibfield{author}{\bibinfo{person}{Yinpeng Dong}, \bibinfo{person}{Fangzhou Liao}, \bibinfo{person}{Tianyu Pang}, \bibinfo{person}{Hang Su}, \bibinfo{person}{Jun Zhu}, \bibinfo{person}{Xiaolin Hu}, {and} \bibinfo{person}{Jianguo Li}.} \bibinfo{year}{2018}\natexlab{}.
\newblock \showarticletitle{Boosting adversarial attacks with momentum}. In \bibinfo{booktitle}{\emph{Proceedings of the IEEE conference on computer vision and pattern recognition}}. \bibinfo{pages}{9185--9193}.
\newblock


\bibitem[Dosovitskiy et~al\mbox{.}(2020)]%
        {dosovitskiy2020image}
\bibfield{author}{\bibinfo{person}{Alexey Dosovitskiy}, \bibinfo{person}{Lucas Beyer}, \bibinfo{person}{Alexander Kolesnikov}, \bibinfo{person}{Dirk Weissenborn}, \bibinfo{person}{Xiaohua Zhai}, \bibinfo{person}{Thomas Unterthiner}, {et~al\mbox{.}}} \bibinfo{year}{2020}\natexlab{}.
\newblock \showarticletitle{An image is worth 16x16 words: Transformers for image recognition at scale}.
\newblock \bibinfo{journal}{\emph{preprint arXiv:2010.11929}} (\bibinfo{year}{2020}).
\newblock


\bibitem[Fort(2022)]%
        {fort2022adversarial}
\bibfield{author}{\bibinfo{person}{Stanislav Fort}.} \bibinfo{year}{2022}\natexlab{}.
\newblock \showarticletitle{Adversarial vulnerability of powerful near out-of-distribution detection}.
\newblock \bibinfo{journal}{\emph{arXiv:2201.07012}} (\bibinfo{year}{2022}).
\newblock


\bibitem[Fursov et~al\mbox{.}(2021)]%
        {fursov2021adversarial}
\bibfield{author}{\bibinfo{person}{Ivan Fursov}, \bibinfo{person}{Matvey Morozov}, \bibinfo{person}{Nina Kaploukhaya}, {et~al\mbox{.}}} \bibinfo{year}{2021}\natexlab{}.
\newblock \showarticletitle{Adversarial attacks on deep models for financial transaction records}. In \bibinfo{booktitle}{\emph{Proceedings of the 27th ACM SIGKDD Conference on Knowledge Discovery \& Data Mining}}.
\newblock


\bibitem[Gatys et~al\mbox{.}(2016)]%
        {style_changes}
\bibfield{author}{\bibinfo{person}{Leon~A. Gatys}, \bibinfo{person}{Alexander~S. Ecker}, {and} \bibinfo{person}{Matthias Bethge}.} \bibinfo{year}{2016}\natexlab{}.
\newblock \showarticletitle{Image Style Transfer Using Convolutional Neural Networks}. In \bibinfo{booktitle}{\emph{2016 IEEE Conference on Computer Vision and Pattern Recognition (CVPR)}}. \bibinfo{pages}{2414--2423}.
\newblock


\bibitem[Ghassemi and Fazl-Ersi(2022)]%
        {ghassemi2022comprehensive}
\bibfield{author}{\bibinfo{person}{Navid Ghassemi} {and} \bibinfo{person}{Ehsan Fazl-Ersi}.} \bibinfo{year}{2022}\natexlab{}.
\newblock \showarticletitle{A Comprehensive Review of Trends, Applications and Challenges In Out-of-Distribution Detection}.
\newblock \bibinfo{journal}{\emph{arXiv preprint arXiv:2209.12935}} (\bibinfo{year}{2022}).
\newblock


\bibitem[Goodfellow et~al\mbox{.}(2014)]%
        {fgsm}
\bibfield{author}{\bibinfo{person}{Ian~J. Goodfellow}, \bibinfo{person}{Jonathon Shlens}, {and} \bibinfo{person}{Christian Szegedy}.} \bibinfo{year}{2014}\natexlab{}.
\newblock \bibinfo{title}{Explaining and Harnessing Adversarial Examples}.
\newblock
\newblock
\urldef\tempurl%
\url{https://doi.org/10.48550/ARXIV.1412.6572}
\showDOI{\tempurl}


\bibitem[Gorbett and Blanchard(2022)]%
        {gorbett2022utilizing}
\bibfield{author}{\bibinfo{person}{Matt Gorbett} {and} \bibinfo{person}{Nathaniel Blanchard}.} \bibinfo{year}{2022}\natexlab{}.
\newblock \showarticletitle{Utilizing network features to detect erroneous inputs}. In \bibinfo{booktitle}{\emph{Proceedings of the IEEE/CVF Winter Conference on Applications of Computer Vision}}. \bibinfo{pages}{34--43}.
\newblock


\bibitem[Gowal et~al\mbox{.}(2018)]%
        {gowal2018effectiveness}
\bibfield{author}{\bibinfo{person}{Sven Gowal}, \bibinfo{person}{Krishnamurthy Dvijotham}, \bibinfo{person}{Robert Stanforth}, \bibinfo{person}{Rudy Bunel}, \bibinfo{person}{Chongli Qin}, {et~al\mbox{.}}} \bibinfo{year}{2018}\natexlab{}.
\newblock \showarticletitle{On the effectiveness of interval bound propagation for training verifiably robust models}.
\newblock \bibinfo{journal}{\emph{preprint arXiv:1810.12715}} (\bibinfo{year}{2018}).
\newblock


\bibitem[He et~al\mbox{.}(2015)]%
        {closed_world_2}
\bibfield{author}{\bibinfo{person}{K. He}, \bibinfo{person}{X. Zhang}, \bibinfo{person}{S. Ren}, {and} \bibinfo{person}{J. Sun}.} \bibinfo{year}{2015}\natexlab{}.
\newblock \showarticletitle{Delving Deep into Rectifiers: Surpassing Human-Level Performance on ImageNet Classification}. In \bibinfo{booktitle}{\emph{2015 IEEE International Conference on Computer Vision (ICCV)}}.
\newblock


\bibitem[He et~al\mbox{.}(2016)]%
        {he2016deep}
\bibfield{author}{\bibinfo{person}{Kaiming He}, \bibinfo{person}{Xiangyu Zhang}, \bibinfo{person}{Shaoqing Ren}, {and} \bibinfo{person}{Jian Sun}.} \bibinfo{year}{2016}\natexlab{}.
\newblock \showarticletitle{Deep residual learning for image recognition}. In \bibinfo{booktitle}{\emph{Proceedings of the IEEE conference on computer vision and pattern recognition}}. \bibinfo{pages}{770--778}.
\newblock


\bibitem[Hein et~al\mbox{.}(2019)]%
        {hein2019relu}
\bibfield{author}{\bibinfo{person}{Matthias Hein}, \bibinfo{person}{Maksym Andriushchenko}, {and} \bibinfo{person}{Julian Bitterwolf}.} \bibinfo{year}{2019}\natexlab{}.
\newblock \showarticletitle{Why relu networks yield high-confidence predictions far away from the training data and how to mitigate the problem}. In \bibinfo{booktitle}{\emph{Proceedings of the IEEE/CVF Conference on Computer Vision and Pattern Recognition}}. \bibinfo{pages}{41--50}.
\newblock


\bibitem[Hendrycks and Dietterich(2019)]%
        {hendrycks2019benchmarking}
\bibfield{author}{\bibinfo{person}{Dan Hendrycks} {and} \bibinfo{person}{Thomas Dietterich}.} \bibinfo{year}{2019}\natexlab{}.
\newblock \showarticletitle{Benchmarking neural network robustness to common corruptions and perturbations}.
\newblock \bibinfo{journal}{\emph{arXiv preprint arXiv:1903.12261}} (\bibinfo{year}{2019}).
\newblock


\bibitem[Hendrycks and Gimpel(2016)]%
        {hendrycks2016baseline}
\bibfield{author}{\bibinfo{person}{Dan Hendrycks} {and} \bibinfo{person}{Kevin Gimpel}.} \bibinfo{year}{2016}\natexlab{}.
\newblock \showarticletitle{A baseline for detecting misclassified and out-of-distribution examples in neural networks}.
\newblock \bibinfo{journal}{\emph{arXiv preprint arXiv:1610.02136}} (\bibinfo{year}{2016}).
\newblock


\bibitem[Hendrycks et~al\mbox{.}(2019a)]%
        {hendrycks2019using_pre-trained}
\bibfield{author}{\bibinfo{person}{Dan Hendrycks}, \bibinfo{person}{Kimin Lee}, {and} \bibinfo{person}{Mantas Mazeika}.} \bibinfo{year}{2019}\natexlab{a}.
\newblock \showarticletitle{Using pre-training can improve model robustness and uncertainty}. In \bibinfo{booktitle}{\emph{International conference on machine learning}}. PMLR, \bibinfo{pages}{2712--2721}.
\newblock


\bibitem[Hendrycks et~al\mbox{.}(2019b)]%
        {outlierexposure}
\bibfield{author}{\bibinfo{person}{Dan Hendrycks}, \bibinfo{person}{Mantas Mazeika}, {and} \bibinfo{person}{Thomas Dietterich}.} \bibinfo{year}{2019}\natexlab{b}.
\newblock \showarticletitle{Deep Anomaly Detection with Outlier Exposure}. In \bibinfo{booktitle}{\emph{International Conference on Learning Representations}}.
\newblock
\urldef\tempurl%
\url{https://openreview.net/forum?id=HyxCxhRcY7}
\showURL{%
\tempurl}


\bibitem[Hendrycks et~al\mbox{.}(2019c)]%
        {hendrycks2019using_self_supervised}
\bibfield{author}{\bibinfo{person}{Dan Hendrycks}, \bibinfo{person}{Mantas Mazeika}, \bibinfo{person}{Saurav Kadavath}, {and} \bibinfo{person}{Dawn Song}.} \bibinfo{year}{2019}\natexlab{c}.
\newblock \showarticletitle{Using self-supervised learning can improve model robustness and uncertainty}.
\newblock \bibinfo{journal}{\emph{Advances in neural information processing systems}}  \bibinfo{volume}{32} (\bibinfo{year}{2019}).
\newblock


\bibitem[Hendrycks et~al\mbox{.}(2022)]%
        {hendrycks2022pixmix}
\bibfield{author}{\bibinfo{person}{Dan Hendrycks}, \bibinfo{person}{Andy Zou}, \bibinfo{person}{Mantas Mazeika}, {et~al\mbox{.}}} \bibinfo{year}{2022}\natexlab{}.
\newblock \showarticletitle{Pixmix: Dreamlike pictures comprehensively improve safety measures}. In \bibinfo{booktitle}{\emph{Proceedings of the IEEE/CVF Conference on Computer Vision and Pattern Recognition}}. \bibinfo{pages}{16783--16792}.
\newblock


\bibitem[Hilal et~al\mbox{.}(2022)]%
        {hilal2022financial}
\bibfield{author}{\bibinfo{person}{Waleed Hilal}, \bibinfo{person}{S~Andrew Gadsden}, {and} \bibinfo{person}{John Yawney}.} \bibinfo{year}{2022}\natexlab{}.
\newblock \showarticletitle{Financial fraud: a review of anomaly detection techniques and recent advances}.
\newblock \bibinfo{journal}{\emph{Expert systems With applications}}  \bibinfo{volume}{193} (\bibinfo{year}{2022}), \bibinfo{pages}{116429}.
\newblock


\bibitem[Huang et~al\mbox{.}(2017)]%
        {huang2017densely}
\bibfield{author}{\bibinfo{person}{Gao Huang}, \bibinfo{person}{Zhuang Liu}, \bibinfo{person}{Laurens Van Der~Maaten}, {and} \bibinfo{person}{Kilian~Q Weinberger}.} \bibinfo{year}{2017}\natexlab{}.
\newblock \showarticletitle{Densely connected convolutional networks}. In \bibinfo{booktitle}{\emph{Proceedings of the IEEE conference on computer vision and pattern recognition}}. \bibinfo{pages}{4700--4708}.
\newblock


\bibitem[{Hull}(1994)]%
        {uspsdataset}
\bibfield{author}{\bibinfo{person}{J.~J. {Hull}}.} \bibinfo{year}{1994}\natexlab{}.
\newblock \showarticletitle{A database for handwritten text recognition research}.
\newblock \bibinfo{journal}{\emph{IEEE Transactions on Pattern Analysis and Machine Intelligence}} \bibinfo{volume}{16}, \bibinfo{number}{5} (\bibinfo{year}{1994}), \bibinfo{pages}{550--554}.
\newblock
\urldef\tempurl%
\url{https://doi.org/10.1109/34.291440}
\showDOI{\tempurl}


\bibitem[Kaur et~al\mbox{.}(2022)]%
        {kaur2022idecode}
\bibfield{author}{\bibinfo{person}{Ramneet Kaur}, \bibinfo{person}{Susmit Jha}, \bibinfo{person}{Anirban Roy}, \bibinfo{person}{Sangdon Park}, {et~al\mbox{.}}} \bibinfo{year}{2022}\natexlab{}.
\newblock \showarticletitle{iDECODe: In-distribution equivariance for conformal out-of-distribution detection}. In \bibinfo{booktitle}{\emph{Proceedings of the AAAI Conference on Artificial Intelligence}}.
\newblock


\bibitem[Kaya et~al\mbox{.}(2022)]%
        {kaya2022generating}
\bibfield{author}{\bibinfo{person}{Yigitcan Kaya}, \bibinfo{person}{Bilal Zafar}, \bibinfo{person}{Sergul Aydore}, \bibinfo{person}{Nathalie Rauschmayr}, {and} \bibinfo{person}{Krishnaram Kenthapadi}.} \bibinfo{year}{2022}\natexlab{}.
\newblock \showarticletitle{Generating distributional adversarial examples to evade statistical detectors}.
\newblock  (\bibinfo{year}{2022}).
\newblock


\bibitem[Khalid et~al\mbox{.}(2022)]%
        {khalid2022rodd}
\bibfield{author}{\bibinfo{person}{Umar Khalid}, \bibinfo{person}{Ashkan Esmaeili}, \bibinfo{person}{Nazmul Karim}, {and} \bibinfo{person}{Nazanin Rahnavard}.} \bibinfo{year}{2022}\natexlab{}.
\newblock \showarticletitle{Rodd: A self-supervised approach for robust out-of-distribution detection}. In \bibinfo{booktitle}{\emph{2022 IEEE/CVF Conference on Computer Vision and Pattern Recognition Workshops (CVPRW)}}. IEEE, \bibinfo{pages}{163--170}.
\newblock


\bibitem[Kim et~al\mbox{.}(2020)]%
        {kim2020adversarial}
\bibfield{author}{\bibinfo{person}{Minseon Kim}, \bibinfo{person}{Jihoon Tack}, {and} \bibinfo{person}{Sung~Ju Hwang}.} \bibinfo{year}{2020}\natexlab{}.
\newblock \showarticletitle{Adversarial self-supervised contrastive learning}.
\newblock \bibinfo{journal}{\emph{Advances in Neural Information Processing Systems}}  \bibinfo{volume}{33} (\bibinfo{year}{2020}), \bibinfo{pages}{2983--2994}.
\newblock


\bibitem[Koh et~al\mbox{.}(2021)]%
        {distributional_shifts}
\bibfield{author}{\bibinfo{person}{Pang~Wei Koh}, \bibinfo{person}{Shiori Sagawa}, \bibinfo{person}{Henrik Marklund}, \bibinfo{person}{Sang~Michael Xie}, \bibinfo{person}{Marvin Zhang}, \bibinfo{person}{Akshay Balsubramani}, {et~al\mbox{.}}} \bibinfo{year}{2021}\natexlab{}.
\newblock \bibinfo{title}{WILDS: A Benchmark of in-the-Wild Distribution Shifts}.
\newblock
\newblock
\showeprint[arxiv]{2012.07421}~[cs.LG]


\bibitem[Kong et~al\mbox{.}(2020)]%
        {kong2020physgan}
\bibfield{author}{\bibinfo{person}{Zelun Kong}, \bibinfo{person}{Junfeng Guo}, \bibinfo{person}{Ang Li}, {and} \bibinfo{person}{Cong Liu}.} \bibinfo{year}{2020}\natexlab{}.
\newblock \showarticletitle{Physgan: Generating physical-world-resilient adversarial examples for autonomous driving}. In \bibinfo{booktitle}{\emph{Proceedings of the IEEE/CVF Conference on Computer Vision and Pattern Recognition}}.
\newblock


\bibitem[Krizhevsky et~al\mbox{.}({[n.\,d.]})]%
        {cifar100}
\bibfield{author}{\bibinfo{person}{Alex Krizhevsky}, \bibinfo{person}{Vinod Nair}, {and} \bibinfo{person}{Geoffrey Hinton}.} \bibinfo{year}{[n.\,d.]}\natexlab{}.
\newblock \showarticletitle{CIFAR-100 (Canadian Institute for Advanced Research)}.
\newblock  (\bibinfo{year}{[n.\,d.]}).
\newblock
\urldef\tempurl%
\url{http://www.cs.toronto.edu/~kriz/cifar.html}
\showURL{%
\tempurl}


\bibitem[Krizhevsky et~al\mbox{.}(2009)]%
        {cifar10}
\bibfield{author}{\bibinfo{person}{Alex Krizhevsky}, \bibinfo{person}{Vinod Nair}, {and} \bibinfo{person}{Geoffrey Hinton}.} \bibinfo{year}{2009}\natexlab{}.
\newblock \showarticletitle{CIFAR-10 (Canadian Institute for Advanced Research)}.
\newblock  (\bibinfo{year}{2009}).
\newblock
\urldef\tempurl%
\url{http://www.cs.toronto.edu/~kriz/cifar.html}
\showURL{%
\tempurl}


\bibitem[Krizhevsky et~al\mbox{.}(2012a)]%
        {closed_world_1}
\bibfield{author}{\bibinfo{person}{Alex Krizhevsky}, \bibinfo{person}{Ilya Sutskever}, {and} \bibinfo{person}{Geoffrey~E Hinton}.} \bibinfo{year}{2012}\natexlab{a}.
\newblock \showarticletitle{ImageNet Classification with Deep Convolutional Neural Networks}. In \bibinfo{booktitle}{\emph{Advances in Neural Information Processing Systems}}, Vol.~\bibinfo{volume}{25}.
\newblock


\bibitem[Krizhevsky et~al\mbox{.}(2012b)]%
        {krizhevsky2012imagenet}
\bibfield{author}{\bibinfo{person}{Alex Krizhevsky}, \bibinfo{person}{Ilya Sutskever}, {and} \bibinfo{person}{Geoffrey~E Hinton}.} \bibinfo{year}{2012}\natexlab{b}.
\newblock \showarticletitle{Imagenet classification with deep convolutional neural networks}.
\newblock \bibinfo{journal}{\emph{Advances in neural information processing systems}}  \bibinfo{volume}{25} (\bibinfo{year}{2012}).
\newblock


\bibitem[Kurakin et~al\mbox{.}(2018)]%
        {kurakin2018adversarial}
\bibfield{author}{\bibinfo{person}{Alexey Kurakin}, \bibinfo{person}{Ian~J Goodfellow}, {and} \bibinfo{person}{Samy Bengio}.} \bibinfo{year}{2018}\natexlab{}.
\newblock \showarticletitle{Adversarial examples in the physical world}.
\newblock In \bibinfo{booktitle}{\emph{Artificial intelligence safety and security}}. \bibinfo{publisher}{Chapman and Hall/CRC}, \bibinfo{pages}{99--112}.
\newblock


\bibitem[Le and Yang(2015)]%
        {Le2015TinyIV}
\bibfield{author}{\bibinfo{person}{Ya Le} {and} \bibinfo{person}{Xuan~S. Yang}.} \bibinfo{year}{2015}\natexlab{}.
\newblock \showarticletitle{Tiny ImageNet Visual Recognition Challenge}.
\newblock
\urldef\tempurl%
\url{https://api.semanticscholar.org/CorpusID:16664790}
\showURL{%
\tempurl}


\bibitem[Lee et~al\mbox{.}(2022)]%
        {lee2022gradient}
\bibfield{author}{\bibinfo{person}{Jinsol Lee}, \bibinfo{person}{Mohit Prabhushankar}, {and} \bibinfo{person}{Ghassan AlRegib}.} \bibinfo{year}{2022}\natexlab{}.
\newblock \showarticletitle{Gradient-based adversarial and out-of-distribution detection}.
\newblock \bibinfo{journal}{\emph{arXiv preprint arXiv:2206.08255}} (\bibinfo{year}{2022}).
\newblock


\bibitem[Lee et~al\mbox{.}(2018)]%
        {lee2018simple}
\bibfield{author}{\bibinfo{person}{Kimin Lee}, \bibinfo{person}{Kibok Lee}, \bibinfo{person}{Honglak Lee}, {and} \bibinfo{person}{Jinwoo Shin}.} \bibinfo{year}{2018}\natexlab{}.
\newblock \showarticletitle{A simple unified framework for detecting out-of-distribution samples and adversarial attacks}.
\newblock \bibinfo{journal}{\emph{Advances in neural information processing systems}}  \bibinfo{volume}{31} (\bibinfo{year}{2018}).
\newblock


\bibitem[Lee et~al\mbox{.}(2020)]%
        {lee2020adversarial}
\bibfield{author}{\bibinfo{person}{Kyungmin Lee}, \bibinfo{person}{Hunmin Yang}, {and} \bibinfo{person}{Se-Yoon Oh}.} \bibinfo{year}{2020}\natexlab{}.
\newblock \showarticletitle{Adversarial training on joint energy based model for robust classification and out-of-distribution detection}. In \bibinfo{booktitle}{\emph{2020 20th ICCAS}}. IEEE, \bibinfo{pages}{17--21}.
\newblock


\bibitem[Li et~al\mbox{.}(2017)]%
        {covariate_2}
\bibfield{author}{\bibinfo{person}{Da Li}, \bibinfo{person}{Yongxin Yang}, \bibinfo{person}{Yi-Zhe Song}, {and} \bibinfo{person}{Timothy~M Hospedales}.} \bibinfo{year}{2017}\natexlab{}.
\newblock \showarticletitle{Deeper, broader and artier domain generalization}. In \bibinfo{booktitle}{\emph{Proceedings of the IEEE international conference on computer vision}}. \bibinfo{pages}{5542--5550}.
\newblock


\bibitem[Liang et~al\mbox{.}(2022)]%
        {liang2022adversarial}
\bibfield{author}{\bibinfo{person}{Hongshuo Liang}, \bibinfo{person}{Erlu He}, \bibinfo{person}{Yangyang Zhao}, \bibinfo{person}{Zhe Jia}, {and} \bibinfo{person}{Hao Li}.} \bibinfo{year}{2022}\natexlab{}.
\newblock \showarticletitle{Adversarial attack and defense: A survey}.
\newblock \bibinfo{journal}{\emph{Electronics}} \bibinfo{volume}{11}, \bibinfo{number}{8} (\bibinfo{year}{2022}), \bibinfo{pages}{1283}.
\newblock


\bibitem[Liang et~al\mbox{.}(2017)]%
        {liang2017enhancing}
\bibfield{author}{\bibinfo{person}{Shiyu Liang}, \bibinfo{person}{Yixuan Li}, {and} \bibinfo{person}{Rayadurgam Srikant}.} \bibinfo{year}{2017}\natexlab{}.
\newblock \showarticletitle{Enhancing the reliability of out-of-distribution image detection in neural networks}.
\newblock \bibinfo{journal}{\emph{arXiv preprint arXiv:1706.02690}} (\bibinfo{year}{2017}).
\newblock


\bibitem[Liu et~al\mbox{.}(2020)]%
        {energy}
\bibfield{author}{\bibinfo{person}{Weitang Liu}, \bibinfo{person}{Xiaoyun Wang}, \bibinfo{person}{John~D. Owens}, {and} \bibinfo{person}{Yixuan Li}.} \bibinfo{year}{2020}\natexlab{}.
\newblock \showarticletitle{Energy-Based out-of-Distribution Detection}. In \bibinfo{booktitle}{\emph{Proceedings of the 34th International Conference on Neural Information Processing Systems}} \emph{(\bibinfo{series}{NIPS'20})}.
\newblock
\showISBNx{9781713829546}


\bibitem[Liu et~al\mbox{.}(2015)]%
        {liu2015deep}
\bibfield{author}{\bibinfo{person}{Ziwei Liu}, \bibinfo{person}{Ping Luo}, \bibinfo{person}{Xiaogang Wang}, {and} \bibinfo{person}{Xiaoou Tang}.} \bibinfo{year}{2015}\natexlab{}.
\newblock \showarticletitle{Deep learning face attributes in the wild}. In \bibinfo{booktitle}{\emph{Proceedings of the IEEE international conference on computer vision}}. \bibinfo{pages}{3730--3738}.
\newblock


\bibitem[Lu et~al\mbox{.}(2023)]%
        {lu2023likelihood}
\bibfield{author}{\bibinfo{person}{Fan Lu}, \bibinfo{person}{Kai Zhu}, \bibinfo{person}{Kecheng Zheng}, \bibinfo{person}{Wei Zhai}, {and} \bibinfo{person}{Yang Cao}.} \bibinfo{year}{2023}\natexlab{}.
\newblock \showarticletitle{Likelihood-Aware Semantic Alignment for Full-Spectrum Out-of-Distribution Detection}.
\newblock \bibinfo{journal}{\emph{arXiv preprint arXiv:2312.01732}} (\bibinfo{year}{2023}).
\newblock


\bibitem[Ma et~al\mbox{.}(2018)]%
        {ma2018characterizing}
\bibfield{author}{\bibinfo{person}{Xingjun Ma}, \bibinfo{person}{Bo Li}, \bibinfo{person}{Yisen Wang}, \bibinfo{person}{Sarah~M Erfani}, \bibinfo{person}{Sudanthi Wijewickrema}, {et~al\mbox{.}}} \bibinfo{year}{2018}\natexlab{}.
\newblock \showarticletitle{Characterizing adversarial subspaces using local intrinsic dimensionality}.
\newblock \bibinfo{journal}{\emph{arXiv preprint arXiv:1801.02613}} (\bibinfo{year}{2018}).
\newblock


\bibitem[Madry et~al\mbox{.}(2017)]%
        {madry2017towards}
\bibfield{author}{\bibinfo{person}{Aleksander Madry}, \bibinfo{person}{Aleksandar Makelov}, \bibinfo{person}{Ludwig Schmidt}, \bibinfo{person}{Dimitris Tsipras}, {and} \bibinfo{person}{Adrian Vladu}.} \bibinfo{year}{2017}\natexlab{}.
\newblock \showarticletitle{Towards deep learning models resistant to adversarial attacks}.
\newblock \bibinfo{journal}{\emph{arXiv preprint arXiv:1706.06083}} (\bibinfo{year}{2017}).
\newblock


\bibitem[Mahdavi and Carvalho(2021)]%
        {open_set_survey}
\bibfield{author}{\bibinfo{person}{A. Mahdavi} {and} \bibinfo{person}{M. Carvalho}.} \bibinfo{year}{2021}\natexlab{}.
\newblock \showarticletitle{A Survey on Open Set Recognition}. In \bibinfo{booktitle}{\emph{2021 IEEE Fourth International Conference on Artificial Intelligence and Knowledge Engineering (AIKE)}}. \bibinfo{pages}{37--44}.
\newblock


\bibitem[Malinin and Gales(2018)]%
        {malinin2018predictive}
\bibfield{author}{\bibinfo{person}{Andrey Malinin} {and} \bibinfo{person}{Mark Gales}.} \bibinfo{year}{2018}\natexlab{}.
\newblock \showarticletitle{Predictive uncertainty estimation via prior networks}.
\newblock \bibinfo{journal}{\emph{Advances in neural information processing systems}}  \bibinfo{volume}{31} (\bibinfo{year}{2018}).
\newblock


\bibitem[Malinin and Gales(2019)]%
        {malinin2019reverse}
\bibfield{author}{\bibinfo{person}{Andrey Malinin} {and} \bibinfo{person}{Mark Gales}.} \bibinfo{year}{2019}\natexlab{}.
\newblock \showarticletitle{Reverse kl-divergence training of prior networks: Improved uncertainty and adversarial robustness}.
\newblock \bibinfo{journal}{\emph{Advances in Neural Information Processing Systems}}  \bibinfo{volume}{32} (\bibinfo{year}{2019}).
\newblock


\bibitem[Megyeri et~al\mbox{.}({[n.\,d.]})]%
        {megyeri2021robust}
\bibfield{author}{\bibinfo{person}{Istv{\'a}n Megyeri}, \bibinfo{person}{Istv{\'a}n Heged{\"u}s}, {and} \bibinfo{person}{M{\'a}rk Jelasity}.} \bibinfo{year}{[n.\,d.]}\natexlab{}.
\newblock \showarticletitle{Robust Classification Combined with Robust out-of-Distribution Detection: An Empirical Analysis}. In \bibinfo{booktitle}{\emph{2021 International Joint Conference on Neural Networks (IJCNN)}}.
\newblock


\bibitem[Meinke et~al\mbox{.}(2022)]%
        {meinke2022provably}
\bibfield{author}{\bibinfo{person}{Alexander Meinke}, \bibinfo{person}{Julian Bitterwolf}, {and} \bibinfo{person}{Matthias Hein}.} \bibinfo{year}{2022}\natexlab{}.
\newblock \showarticletitle{Provably Adversarially Robust Detection of Out-of-distribution Data (almost) for free}.
\newblock \bibinfo{journal}{\emph{Advances in Neural Information Processing Systems}}  \bibinfo{volume}{35} (\bibinfo{year}{2022}), \bibinfo{pages}{30167--30180}.
\newblock


\bibitem[Meinke and Hein(2019)]%
        {meinke2019towards}
\bibfield{author}{\bibinfo{person}{Alexander Meinke} {and} \bibinfo{person}{Matthias Hein}.} \bibinfo{year}{2019}\natexlab{}.
\newblock \showarticletitle{Towards neural networks that provably know when they don't know}. In \bibinfo{booktitle}{\emph{International Conference on Learning Representations}}.
\newblock


\bibitem[Moosavi-Dezfooli et~al\mbox{.}(2016)]%
        {moosavi2016deepfool}
\bibfield{author}{\bibinfo{person}{Seyed-Mohsen Moosavi-Dezfooli}, \bibinfo{person}{Alhussein Fawzi}, {and} \bibinfo{person}{Pascal Frossard}.} \bibinfo{year}{2016}\natexlab{}.
\newblock \showarticletitle{Deepfool: a simple and accurate method to fool deep neural networks}. In \bibinfo{booktitle}{\emph{Proceedings of the IEEE conference on computer vision and pattern recognition}}.
\newblock


\bibitem[Naseer et~al\mbox{.}(2020)]%
        {naseer2020self}
\bibfield{author}{\bibinfo{person}{Muzammal Naseer}, \bibinfo{person}{Salman Khan}, \bibinfo{person}{Munawar Hayat}, {et~al\mbox{.}}} \bibinfo{year}{2020}\natexlab{}.
\newblock \showarticletitle{A self-supervised approach for adversarial robustness}. In \bibinfo{booktitle}{\emph{Proceedings of the IEEE/CVF Conference on Computer Vision and Pattern Recognition}}. \bibinfo{pages}{262--271}.
\newblock


\bibitem[Netzer et~al\mbox{.}(2011)]%
        {SVHN}
\bibfield{author}{\bibinfo{person}{Yuval Netzer}, \bibinfo{person}{Tao Wang}, \bibinfo{person}{Adam Coates}, \bibinfo{person}{Alessandro Bissacco}, {et~al\mbox{.}}} \bibinfo{year}{2011}\natexlab{}.
\newblock \showarticletitle{Reading Digits in Natural Images with Unsupervised Feature Learning}. In \bibinfo{booktitle}{\emph{NIPS Workshop on Deep Learning and Unsupervised Feature Learning 2011}}.
\newblock


\bibitem[Nguyen et~al\mbox{.}(2015)]%
        {nguyen2015deep}
\bibfield{author}{\bibinfo{person}{Anh Nguyen}, \bibinfo{person}{Jason Yosinski}, {and} \bibinfo{person}{Jeff Clune}.} \bibinfo{year}{2015}\natexlab{}.
\newblock \showarticletitle{Deep neural networks are easily fooled: High confidence predictions for unrecognizable images}. In \bibinfo{booktitle}{\emph{Proceedings of the IEEE conference on computer vision and pattern recognition}}. \bibinfo{pages}{427--436}.
\newblock


\bibitem[Nijkamp et~al\mbox{.}(2020)]%
        {nijkamp2020anatomy}
\bibfield{author}{\bibinfo{person}{Erik Nijkamp}, \bibinfo{person}{Mitch Hill}, \bibinfo{person}{Tian Han}, \bibinfo{person}{Song-Chun Zhu}, {and} \bibinfo{person}{Ying~Nian Wu}.} \bibinfo{year}{2020}\natexlab{}.
\newblock \showarticletitle{On the anatomy of mcmc-based maximum likelihood learning of energy-based models}. In \bibinfo{booktitle}{\emph{Proceedings of the AAAI Conference on Artificial Intelligence}}.
\newblock


\bibitem[Nitsch et~al\mbox{.}(2021)]%
        {nitsch2021out}
\bibfield{author}{\bibinfo{person}{Julia Nitsch}, \bibinfo{person}{Masha Itkina}, \bibinfo{person}{Ransalu Senanayake}, \bibinfo{person}{Juan Nieto}, {et~al\mbox{.}}} \bibinfo{year}{2021}\natexlab{}.
\newblock \showarticletitle{Out-of-distribution detection for automotive perception}. In \bibinfo{booktitle}{\emph{2021 IEEE International Intelligent Transportation Systems Conference (ITSC)}}. IEEE, \bibinfo{pages}{2938--2943}.
\newblock


\bibitem[(NTSB)(2016)]%
        {NTSB2016TeslaCrash}
\bibfield{author}{\bibinfo{person}{National Transportation Safety~Board (NTSB)}.} \bibinfo{year}{2016}\natexlab{}.
\newblock \bibinfo{title}{Collision Between a Car Operating With Automated Vehicle Control Systems and a Tractor-Semitrailer Truck Near Williston, Florida May 7, 2016}.
\newblock
\newblock


\bibitem[Ojaswee et~al\mbox{.}(2023)]%
        {ojaswee2023benchmarking}
\bibfield{author}{\bibinfo{person}{Ojaswee Ojaswee}, \bibinfo{person}{Akshay Agarwal}, {and} \bibinfo{person}{Nalini Ratha}.} \bibinfo{year}{2023}\natexlab{}.
\newblock \showarticletitle{Benchmarking Image Classifiers for Physical Out-of-Distribution Examples Detection}. In \bibinfo{booktitle}{\emph{Proceedings of the IEEE/CVF International Conference on Computer Vision}}.
\newblock


\bibitem[Osada et~al\mbox{.}(2023)]%
        {osada2023out}
\bibfield{author}{\bibinfo{person}{Genki Osada}, \bibinfo{person}{Tsubasa Takahashi}, \bibinfo{person}{Budrul Ahsan}, {and} \bibinfo{person}{Takashi Nishide}.} \bibinfo{year}{2023}\natexlab{}.
\newblock \showarticletitle{Out-of-Distribution Detection with Reconstruction Error and Typicality-based Penalty}. In \bibinfo{booktitle}{\emph{Proceedings of the IEEE/CVF Winter Conference on Applications of Computer Vision}}. \bibinfo{pages}{5551--5563}.
\newblock


\bibitem[Papernot et~al\mbox{.}(2016)]%
        {papernot2016distillation}
\bibfield{author}{\bibinfo{person}{Nicolas Papernot}, \bibinfo{person}{Patrick McDaniel}, \bibinfo{person}{Xi Wu}, \bibinfo{person}{Somesh Jha}, {and} \bibinfo{person}{Ananthram Swami}.} \bibinfo{year}{2016}\natexlab{}.
\newblock \showarticletitle{Distillation as a defense to adversarial perturbations against deep neural networks}. In \bibinfo{booktitle}{\emph{2016 IEEE symposium on security and privacy (SP)}}. IEEE.
\newblock


\bibitem[Pinto et~al\mbox{.}(2022)]%
        {pinto2022using}
\bibfield{author}{\bibinfo{person}{Francesco Pinto}, \bibinfo{person}{Harry Yang}, \bibinfo{person}{Ser~Nam Lim}, \bibinfo{person}{Philip Torr}, {et~al\mbox{.}}} \bibinfo{year}{2022}\natexlab{}.
\newblock \showarticletitle{Using mixup as a regularizer can surprisingly improve accuracy \& out-of-distribution robustness}.
\newblock \bibinfo{journal}{\emph{Advances in Neural Information Processing Systems}} (\bibinfo{year}{2022}).
\newblock


\bibitem[Qiu and Miikkulainen(2022)]%
        {qiu2022detecting}
\bibfield{author}{\bibinfo{person}{Xin Qiu} {and} \bibinfo{person}{Risto Miikkulainen}.} \bibinfo{year}{2022}\natexlab{}.
\newblock \showarticletitle{Detecting misclassification errors in neural networks with a gaussian process model}. In \bibinfo{booktitle}{\emph{Proceedings of the AAAI Conference on Artificial Intelligence}}, Vol.~\bibinfo{volume}{36}. \bibinfo{pages}{8017--8027}.
\newblock


\bibitem[Quinonero-Candela et~al\mbox{.}(2008)]%
        {quinonero2008dataset}
\bibfield{author}{\bibinfo{person}{Joaquin Quinonero-Candela}, \bibinfo{person}{Masashi Sugiyama}, \bibinfo{person}{Anton Schwaighofer}, {and} \bibinfo{person}{Neil~D Lawrence}.} \bibinfo{year}{2008}\natexlab{}.
\newblock \bibinfo{booktitle}{\emph{Dataset shift in machine learning}}.
\newblock \bibinfo{publisher}{Mit Press}.
\newblock


\bibitem[Ratzlaff and Fuxin(2019)]%
        {ratzlaff2019hypergan}
\bibfield{author}{\bibinfo{person}{Neale Ratzlaff} {and} \bibinfo{person}{Li Fuxin}.} \bibinfo{year}{2019}\natexlab{}.
\newblock \showarticletitle{Hypergan: A generative model for diverse, performant neural networks}. In \bibinfo{booktitle}{\emph{International Conference on Machine Learning}}. PMLR, \bibinfo{pages}{5361--5369}.
\newblock


\bibitem[Ren et~al\mbox{.}(2021)]%
        {ren2021simple}
\bibfield{author}{\bibinfo{person}{Jie Ren}, \bibinfo{person}{Stanislav Fort}, \bibinfo{person}{Jeremiah Liu}, \bibinfo{person}{Abhijit~Guha Roy}, \bibinfo{person}{Shreyas Padhy}, {and} \bibinfo{person}{Balaji Lakshminarayanan}.} \bibinfo{year}{2021}\natexlab{}.
\newblock \showarticletitle{A simple fix to mahalanobis distance for improving near-ood detection}.
\newblock \bibinfo{journal}{\emph{arXiv preprint arXiv:2106.09022}} (\bibinfo{year}{2021}).
\newblock


\bibitem[Salehi et~al\mbox{.}(2021)]%
        {salehi2021unified}
\bibfield{author}{\bibinfo{person}{Mohammadreza Salehi}, \bibinfo{person}{Hossein Mirzaei}, \bibinfo{person}{Dan Hendrycks}, {et~al\mbox{.}}} \bibinfo{year}{2021}\natexlab{}.
\newblock \showarticletitle{A unified survey on anomaly, novelty, open-set, and out-of-distribution detection: Solutions and future challenges}.
\newblock \bibinfo{journal}{\emph{arXiv preprint arXiv:2110.14051}} (\bibinfo{year}{2021}).
\newblock


\bibitem[Saltzer and Schroeder(1975)]%
        {design_principles}
\bibfield{author}{\bibinfo{person}{J.H. Saltzer} {and} \bibinfo{person}{M.D. Schroeder}.} \bibinfo{year}{1975}\natexlab{}.
\newblock \showarticletitle{The protection of information in computer systems}.
\newblock \bibinfo{journal}{\emph{Proc. IEEE}} (\bibinfo{year}{1975}).
\newblock


\bibitem[Sehwag et~al\mbox{.}(2019)]%
        {sehwag2019analyzing}
\bibfield{author}{\bibinfo{person}{Vikash Sehwag}, \bibinfo{person}{Arjun~Nitin Bhagoji}, \bibinfo{person}{Liwei Song}, \bibinfo{person}{Chawin Sitawarin}, {et~al\mbox{.}}} \bibinfo{year}{2019}\natexlab{}.
\newblock \showarticletitle{Analyzing the robustness of open-world machine learning}. In \bibinfo{booktitle}{\emph{Proceedings of the 12th ACM Workshop on Artificial Intelligence and Security}}. \bibinfo{pages}{105--116}.
\newblock


\bibitem[Sensoy et~al\mbox{.}(2020)]%
        {sensoy2020uncertainty}
\bibfield{author}{\bibinfo{person}{Murat Sensoy}, \bibinfo{person}{Lance Kaplan}, \bibinfo{person}{Federico Cerutti}, {and} \bibinfo{person}{Maryam Saleki}.} \bibinfo{year}{2020}\natexlab{}.
\newblock \showarticletitle{Uncertainty-aware deep classifiers using generative models}. In \bibinfo{booktitle}{\emph{Proceedings of the AAAI conference on artificial intelligence}}, Vol.~\bibinfo{volume}{34}. \bibinfo{pages}{5620--5627}.
\newblock


\bibitem[Shen et~al\mbox{.}(2021)]%
        {generalization_survey}
\bibfield{author}{\bibinfo{person}{Zheyan Shen}, \bibinfo{person}{Jiashuo Liu}, \bibinfo{person}{Yue He}, \bibinfo{person}{Xingxuan Zhang}, \bibinfo{person}{Renzhe Xu}, \bibinfo{person}{Han Yu}, {and} \bibinfo{person}{Peng Cui}.} \bibinfo{year}{2021}\natexlab{}.
\newblock \bibinfo{title}{Towards Out-Of-Distribution Generalization: A Survey}.
\newblock
\newblock
\urldef\tempurl%
\url{https://doi.org/10.48550/ARXIV.2108.13624}
\showDOI{\tempurl}


\bibitem[Silva and Najafirad(2020)]%
        {silva2020opportunities}
\bibfield{author}{\bibinfo{person}{Samuel~Henrique Silva} {and} \bibinfo{person}{Peyman Najafirad}.} \bibinfo{year}{2020}\natexlab{}.
\newblock \showarticletitle{Opportunities and challenges in deep learning adversarial robustness: A survey}.
\newblock \bibinfo{journal}{\emph{arXiv preprint arXiv:2007.00753}} (\bibinfo{year}{2020}).
\newblock


\bibitem[Sun et~al\mbox{.}(2021)]%
        {react}
\bibfield{author}{\bibinfo{person}{Yiyou Sun}, \bibinfo{person}{Chuan Guo}, {and} \bibinfo{person}{Yixuan Li}.} \bibinfo{year}{2021}\natexlab{}.
\newblock \showarticletitle{ReAct: Out-of-distribution Detection With Rectified Activations}. In \bibinfo{booktitle}{\emph{Advances in Neural Information Processing Systems}}.
\newblock


\bibitem[Sun and Li(2022)]%
        {dice}
\bibfield{author}{\bibinfo{person}{Yiyou Sun} {and} \bibinfo{person}{Yixuan Li}.} \bibinfo{year}{2022}\natexlab{}.
\newblock \showarticletitle{DICE: Leveraging Sparsification for Out-of-Distribution Detection}. In \bibinfo{booktitle}{\emph{Proceedings of European Conference on Computer Vision}}.
\newblock


\bibitem[Szegedy et~al\mbox{.}(2013)]%
        {szegedy2013intriguing}
\bibfield{author}{\bibinfo{person}{Christian Szegedy}, \bibinfo{person}{Wojciech Zaremba}, \bibinfo{person}{Ilya Sutskever}, \bibinfo{person}{Joan Bruna}, \bibinfo{person}{Dumitru Erhan}, \bibinfo{person}{Ian Goodfellow}, {and} \bibinfo{person}{Rob Fergus}.} \bibinfo{year}{2013}\natexlab{}.
\newblock \showarticletitle{Intriguing properties of neural networks}.
\newblock \bibinfo{journal}{\emph{arXiv preprint arXiv:1312.6199}} (\bibinfo{year}{2013}).
\newblock


\bibitem[Wang({[n.\,d.]})]%
        {wang2020high}
\bibfield{author}{\bibinfo{person}{Tao Wang}.} \bibinfo{year}{[n.\,d.]}\natexlab{}.
\newblock \showarticletitle{High precision open-world website fingerprinting}. In \bibinfo{booktitle}{\emph{2020 IEEE Symposium on Security and Privacy}}.
\newblock


\bibitem[Wei et~al\mbox{.}(2022)]%
        {logitnormalization}
\bibfield{author}{\bibinfo{person}{Hongxin Wei}, \bibinfo{person}{Renchunzi Xie}, \bibinfo{person}{Hao Cheng}, \bibinfo{person}{Lei Feng}, \bibinfo{person}{Bo An}, {and} \bibinfo{person}{Yixuan Li}.} \bibinfo{year}{2022}\natexlab{}.
\newblock \bibinfo{title}{Mitigating Neural Network Overconfidence with Logit Normalization}.
\newblock
\newblock
\urldef\tempurl%
\url{https://doi.org/10.48550/ARXIV.2205.09310}
\showDOI{\tempurl}


\bibitem[Xiao et~al\mbox{.}(2017)]%
        {xiao2017fashion}
\bibfield{author}{\bibinfo{person}{Han Xiao}, \bibinfo{person}{Kashif Rasul}, {and} \bibinfo{person}{Roland Vollgraf}.} \bibinfo{year}{2017}\natexlab{}.
\newblock \showarticletitle{Fashion-mnist: a novel image dataset for benchmarking machine learning algorithms}.
\newblock \bibinfo{journal}{\emph{arXiv preprint arXiv:1708.07747}} (\bibinfo{year}{2017}).
\newblock


\bibitem[Xu et~al\mbox{.}(2015)]%
        {xu2015turkergaze}
\bibfield{author}{\bibinfo{person}{Pingmei Xu}, \bibinfo{person}{Krista~A Ehinger}, \bibinfo{person}{Yinda Zhang}, \bibinfo{person}{Adam Finkelstein}, \bibinfo{person}{Sanjeev~R Kulkarni}, {and} \bibinfo{person}{Jianxiong Xiao}.} \bibinfo{year}{2015}\natexlab{}.
\newblock \showarticletitle{Turkergaze: Crowdsourcing saliency with webcam based eye tracking}.
\newblock \bibinfo{journal}{\emph{arXiv preprint arXiv:1504.06755}} (\bibinfo{year}{2015}).
\newblock


\bibitem[Yang et~al\mbox{.}(2021a)]%
        {yang2021semantically}
\bibfield{author}{\bibinfo{person}{Jingkang Yang}, \bibinfo{person}{Haoqi Wang}, \bibinfo{person}{Litong Feng}, \bibinfo{person}{Xiaopeng Yan}, \bibinfo{person}{Huabin Zheng}, {et~al\mbox{.}}} \bibinfo{year}{2021}\natexlab{a}.
\newblock \showarticletitle{Semantically coherent out-of-distribution detection}. In \bibinfo{booktitle}{\emph{Proceedings of the IEEE/CVF International Conference on Computer Vision}}. \bibinfo{pages}{8301--8309}.
\newblock


\bibitem[Yang et~al\mbox{.}(2022)]%
        {yang2022openood}
\bibfield{author}{\bibinfo{person}{Jingkang Yang}, \bibinfo{person}{Pengyun Wang}, \bibinfo{person}{Dejian Zou}, \bibinfo{person}{Zitang Zhou}, \bibinfo{person}{Kunyuan Ding}, {et~al\mbox{.}}} \bibinfo{year}{2022}\natexlab{}.
\newblock \showarticletitle{Openood: Benchmarking generalized out-of-distribution detection}.
\newblock \bibinfo{journal}{\emph{Advances in Neural Information Processing Systems}}  \bibinfo{volume}{35} (\bibinfo{year}{2022}), \bibinfo{pages}{32598--32611}.
\newblock


\bibitem[Yang et~al\mbox{.}(2021b)]%
        {detection_survey}
\bibfield{author}{\bibinfo{person}{Jingkang Yang}, \bibinfo{person}{Kaiyang Zhou}, \bibinfo{person}{Yixuan Li}, {and} \bibinfo{person}{Ziwei Liu}.} \bibinfo{year}{2021}\natexlab{b}.
\newblock \bibinfo{title}{Generalized Out-of-Distribution Detection: A Survey}.
\newblock
\newblock
\urldef\tempurl%
\url{https://doi.org/10.48550/ARXIV.2110.11334}
\showDOI{\tempurl}


\bibitem[Yang et~al\mbox{.}(2023)]%
        {yang2023full}
\bibfield{author}{\bibinfo{person}{Jingkang Yang}, \bibinfo{person}{Kaiyang Zhou}, {and} \bibinfo{person}{Ziwei Liu}.} \bibinfo{year}{2023}\natexlab{}.
\newblock \showarticletitle{Full-spectrum out-of-distribution detection}.
\newblock \bibinfo{journal}{\emph{International Journal of Computer Vision}} \bibinfo{volume}{131}, \bibinfo{number}{10} (\bibinfo{year}{2023}), \bibinfo{pages}{2607--2622}.
\newblock


\bibitem[Yin et~al\mbox{.}(2022)]%
        {yin2022learning}
\bibfield{author}{\bibinfo{person}{Xuwang Yin}, \bibinfo{person}{Shiying Li}, {and} \bibinfo{person}{Gustavo~K Rohde}.} \bibinfo{year}{2022}\natexlab{}.
\newblock \showarticletitle{Learning energy-based models with adversarial training}. In \bibinfo{booktitle}{\emph{European Conference on Computer Vision}}. Springer, \bibinfo{pages}{209--226}.
\newblock


\bibitem[Yoon et~al\mbox{.}(2022)]%
        {yoon2022evaluating}
\bibfield{author}{\bibinfo{person}{Sangwoong Yoon}, \bibinfo{person}{Jinwon Choi}, \bibinfo{person}{Yonghyeon Lee}, \bibinfo{person}{Yung-Kyun Noh}, {and} \bibinfo{person}{Frank~Chongwoo Park}.} \bibinfo{year}{2022}\natexlab{}.
\newblock \showarticletitle{Evaluating out-of-distribution detectors through adversarial generation of outliers}.
\newblock \bibinfo{journal}{\emph{arXiv preprint arXiv:2208.10940}} (\bibinfo{year}{2022}).
\newblock


\bibitem[Yu et~al\mbox{.}(2015)]%
        {yu2015lsun}
\bibfield{author}{\bibinfo{person}{Fisher Yu}, \bibinfo{person}{Ari Seff}, \bibinfo{person}{Yinda Zhang}, \bibinfo{person}{Shuran Song}, \bibinfo{person}{Thomas Funkhouser}, {and} \bibinfo{person}{Jianxiong Xiao}.} \bibinfo{year}{2015}\natexlab{}.
\newblock \showarticletitle{Lsun: Construction of a large-scale image dataset using deep learning with humans in the loop}.
\newblock \bibinfo{journal}{\emph{arXiv preprint arXiv:1506.03365}} (\bibinfo{year}{2015}).
\newblock


\bibitem[Yu and Aizawa(2019)]%
        {yu2019unsupervised}
\bibfield{author}{\bibinfo{person}{Qing Yu} {and} \bibinfo{person}{Kiyoharu Aizawa}.} \bibinfo{year}{2019}\natexlab{}.
\newblock \showarticletitle{Unsupervised out-of-distribution detection by maximum classifier discrepancy}. In \bibinfo{booktitle}{\emph{Proceedings of the IEEE/CVF international conference on computer vision}}. \bibinfo{pages}{9518--9526}.
\newblock


\bibitem[Zagoruyko and Komodakis(2016)]%
        {zagoruyko2016wide}
\bibfield{author}{\bibinfo{person}{Sergey Zagoruyko} {and} \bibinfo{person}{Nikos Komodakis}.} \bibinfo{year}{2016}\natexlab{}.
\newblock \showarticletitle{Wide residual networks}.
\newblock \bibinfo{journal}{\emph{arXiv preprint arXiv:1605.07146}} (\bibinfo{year}{2016}).
\newblock


\bibitem[Zhang et~al\mbox{.}(2017)]%
        {zhang2017mixup}
\bibfield{author}{\bibinfo{person}{Hongyi Zhang}, \bibinfo{person}{Moustapha Cisse}, \bibinfo{person}{Yann~N Dauphin}, {and} \bibinfo{person}{David Lopez-Paz}.} \bibinfo{year}{2017}\natexlab{}.
\newblock \showarticletitle{mixup: Beyond empirical risk minimization}.
\newblock \bibinfo{journal}{\emph{arXiv preprint arXiv:1710.09412}} (\bibinfo{year}{2017}).
\newblock


\bibitem[Zhang et~al\mbox{.}(2023a)]%
        {zhang2023mixture}
\bibfield{author}{\bibinfo{person}{Jingyang Zhang}, \bibinfo{person}{Nathan Inkawhich}, \bibinfo{person}{Randolph Linderman}, \bibinfo{person}{Yiran Chen}, {and} \bibinfo{person}{Hai Li}.} \bibinfo{year}{2023}\natexlab{a}.
\newblock \showarticletitle{Mixture outlier exposure: Towards out-of-distribution detection in fine-grained environments}. In \bibinfo{booktitle}{\emph{Proceedings of the IEEE/CVF Winter Conference on Applications of Computer Vision}}. \bibinfo{pages}{5531--5540}.
\newblock


\bibitem[Zhang et~al\mbox{.}(2023b)]%
        {zhang2023openood}
\bibfield{author}{\bibinfo{person}{Jingyang Zhang}, \bibinfo{person}{Jingkang Yang}, \bibinfo{person}{Pengyun Wang}, \bibinfo{person}{Haoqi Wang}, \bibinfo{person}{Yueqian Lin}, \bibinfo{person}{Haoran Zhang}, \bibinfo{person}{Yiyou Sun}, {et~al\mbox{.}}} \bibinfo{year}{2023}\natexlab{b}.
\newblock \showarticletitle{OpenOOD v1.5: Enhanced Benchmark for Out-of-Distribution Detection}.
\newblock \bibinfo{journal}{\emph{arXiv preprint arXiv:2306.09301}} (\bibinfo{year}{2023}).
\newblock


\bibitem[Zhou et~al\mbox{.}(2017)]%
        {zhou2017places}
\bibfield{author}{\bibinfo{person}{Bolei Zhou}, \bibinfo{person}{Agata Lapedriza}, \bibinfo{person}{Aditya Khosla}, \bibinfo{person}{Aude Oliva}, {and} \bibinfo{person}{Antonio Torralba}.} \bibinfo{year}{2017}\natexlab{}.
\newblock \showarticletitle{Places: A 10 million image database for scene recognition}.
\newblock \bibinfo{journal}{\emph{IEEE transactions on pattern analysis and machine intelligence}} \bibinfo{volume}{40}, \bibinfo{number}{6} (\bibinfo{year}{2017}), \bibinfo{pages}{1452--1464}.
\newblock


\bibitem[Zhou et~al\mbox{.}(2023)]%
        {domain_generalisation_survey}
\bibfield{author}{\bibinfo{person}{Kaiyang Zhou}, \bibinfo{person}{Ziwei Liu}, \bibinfo{person}{Yu Qiao}, \bibinfo{person}{Tao Xiang}, {and} \bibinfo{person}{Chen~Change Loy}.} \bibinfo{year}{2023}\natexlab{}.
\newblock \showarticletitle{Domain Generalization: A Survey}.
\newblock \bibinfo{journal}{\emph{IEEE Transactions on Pattern Analysis and Machine Intelligence}} \bibinfo{volume}{45}, \bibinfo{number}{4} (\bibinfo{year}{2023}), \bibinfo{pages}{4396--4415}.
\newblock


\end{thebibliography}
% \appendix

\end{document}